\documentclass[a4paper,fleqn]{cas-sc}

\usepackage[numbers]{natbib}
\usepackage{dirtytalk}
\usepackage{tabularx}
\usepackage{subcaption}
\usepackage{gensymb}
\usepackage[abs]{overpic}
\usepackage{pict2e}
\usepackage{stfloats}
\usepackage{hyperref}

\newcommand{\mb}[1]{\mathbf{#1}}

\def\tsc#1{\csdef{#1}{\textsc{\lowercase{#1}}\xspace}}
\tsc{WGM}
\tsc{QE}

\begin{document}
\let\WriteBookmarks\relax
\def\floatpagepagefraction{1}
\def\textpagefraction{.001}

\shorttitle{}    

\shortauthors{}  

\title[mode=title]{Interaction Forces and Internal Loads in Parallel Manipulators with Actuation Redundancy}
\tnotemark[1]
\tnotetext[1]{\textcopyright~2026. This manuscript version is made available
  under the CC-BY-NC-ND 4.0 license
  \url{https://creativecommons.org/licenses/by-nc-nd/4.0/}. The final published version is available at \url{https://doi.org/10.1016/j.mechmachtheory.2026.106573}.}



%

\author[1]{Joshua Flight}[orcid=0009-0000-1794-0646]

\cormark[1]


\ead{joshua.flight.1@ulaval.ca}



\affiliation[1]{organization={Département de génie mécanique et de génie industriel, Université Laval},
            addressline={1065 avenue de la Médecine}, 
            city={Québec},
            postcode={G1V 0A6}, 
            state={Québec},
            country={Canada}}

\author[1]{Clément Gosselin}[orcid=0000-0001-7422-4515]


\ead{clement.gosselin@gmc.ulaval.ca}




\cortext[1]{Corresponding author}



\begin{abstract}
This paper discusses null-space wrench components in parallel manipulators. We examine the adaptation of the two most common characterizations of these components in grasp-like systems, namely, interaction forces and internal loads, to parallel manipulators with actuation redundancy. We identify critical oversights in the existing literature on the subject, resolve ambiguities related to the definitions of interaction forces and internal loads, and provide explicit methods for synthesizing equilibrating and manipulating joint torque vectors. Case studies are also provided to justify the validity of our novel methods and correct erroneous results reported in the literature. 
\end{abstract}





\maketitle


\section{Introduction}
Mechanical systems in which a single object is manipulated or supported by multiple actuated kinematic chains are becoming increasingly widespread in robotics. Prominent examples include multifingered grippers, legged robots, cooperating manipulators, and parallel manipulators. 

\begin{figure}
\centering
    \subfloat[Example of a multifingered gripper that applies pure forces to the handled object. Adapted from \cite{Kumar1988}.]{%
        \includegraphics[width=.32\linewidth]{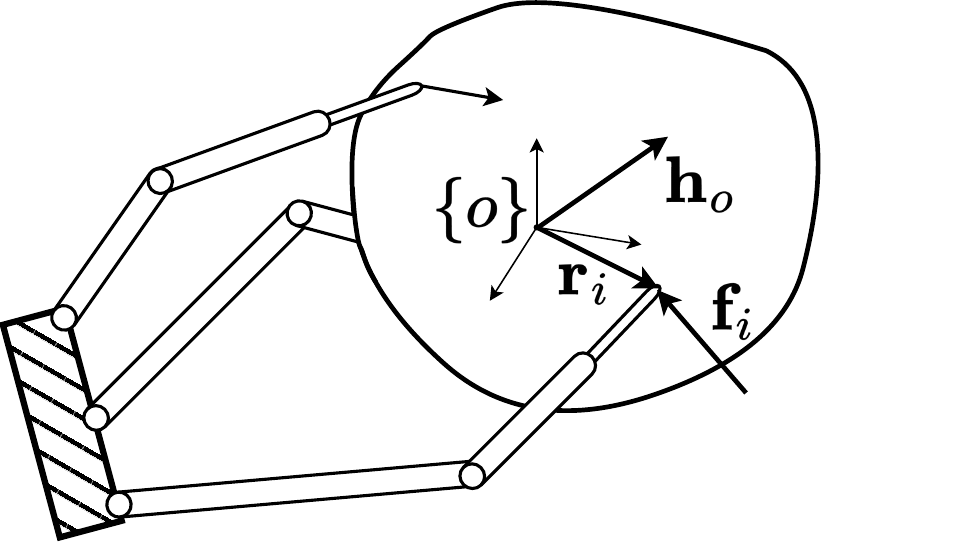}%
        \label{subfig:GripperExample}%
    } \quad
    \subfloat[Example of a legged robot that applies pure forces to the ground.  Adapted from \cite{Gor2013}.]{%
        \includegraphics[width=.32\linewidth]{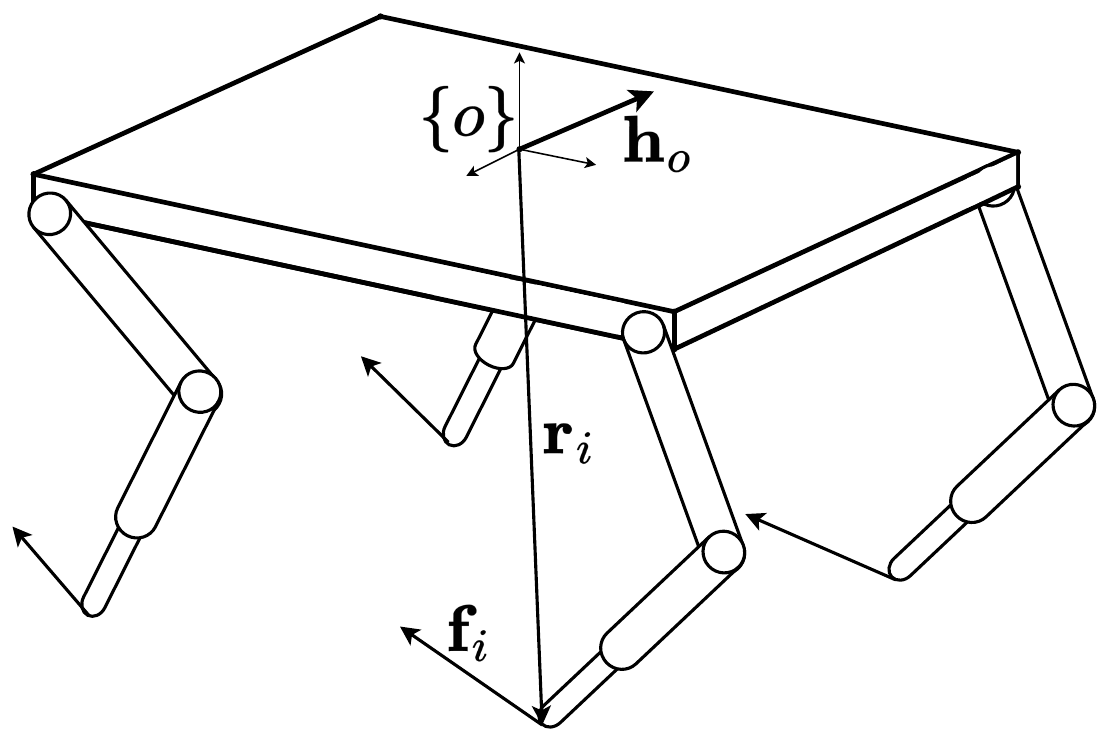}%
        \label{subfig:LeggedRobotExample}%
    } \
    \subfloat[Example of cooperating manipulators each applying a pure force $\mb{f}_i$ and a pure torque $\mb{t}_i$ to the object.]{%
        \includegraphics[width=.35\linewidth]{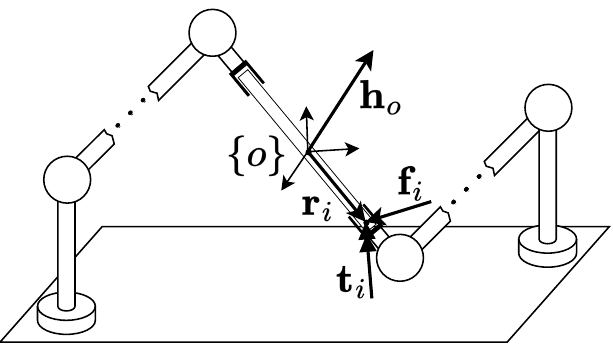}%
        \label{subfig:CoopManipsExample}%
    }
    \caption{Examples of grasp-like systems \cite{Flight2026, Kumar1988, Gor2013}.}
    \label{fig:Examples}
\end{figure}

For multifingered grippers, legged robots and cooperating manipulators, the wrench equilibrium equation is typically formulated using the \emph{grasp matrix} \cite{prattichizzo2008grasping}:

\begin{equation}
    \mb{h}_o = \mb{G} \mb{h}.
    \label{eq:GraspStatics}
\end{equation}
Here, $\mb{h}_o = \begin{bmatrix}
    \mb{f}_o^T & \mb{t}_o^T
\end{bmatrix}^T$ is the resultant wrench applied to the rigid object, comprising a force $\mb{f}_o$ and a torque $\mb{t}_o$. The vector $\mb{h} = \begin{bmatrix}
    \mb{h}_1^T & \ldots  & \mb{h}_n^T
\end{bmatrix}^T$ stacks the $n$ local wrenches applied to the object by the kinematic chains at locations defined by vectors $\mb{r}_i$ for $i = 1, ..., n$ (see Fig. \ref{fig:Examples}), and the grasp matrix $\mb{G}$ maps these local wrenches to the resultant wrench. The nature of vector $\mb{h}$ varies by application. Typically, it defines contact forces in grasping, ground reaction wrenches in legged locomotion, or wrenches applied to an object by the end-effectors of cooperating manipulators. We use the term \emph{grasp-like} systems to refer to these types of robots.


In contrast, parallel manipulators require a distinct formulation. Their wrench equilibrium is not defined by aggregating Cartesian applied wrenches using the grasp matrix, but rather by mapping actuated joint torques to the end-effector wrench. This wrench mapping is governed by the kinematic Jacobians which relate the actuated joint velocities to the end-effector Cartesian velocities. By invoking the principle of virtual work (or static-kinematic duality), we obtain an equivalent mapping between the joint efforts and the end-effector wrench that is fundamentally different from (\ref{eq:GraspStatics}).

While both classes of systems adhere to the same underlying physical principles, this structural difference is non-trivial. Methods developed for grasp-like systems cannot be blindly applied to parallel manipulators; they must be adapted to account for the implicit nature of the Jacobian-based wrench mapping. This discrepancy becomes particularly important when analyzing redundancy. 

\begin{tabular}{l p{0.75\linewidth}}
    \hline
    \textbf{Nomenclature} & \\
    \hline
    $n$ & Number of legs in the parallel manipulator. \\
    $q$ & Total number of actuators in the manipulator. \\
    $p$ & Dimension of the task space (3 for planar manipulators, 6 for spatial manipulators). \\
    $\mb{x}_o$  & Pose of the end-effector. \\
    $\dot{\mb{x}}_o$  & Instantaneous twist of the end-effector. \\
    $\ddot{\mb{x}}_o$  & Instantaneous acceleration of the end-effector. \\
    $\boldsymbol{\theta}$  & Joint position vector. \\
    $\dot{\boldsymbol{\theta}}$  & Joint velocity vector. \\
    $\mb{J}$, $\mb{K}$  & Kinematic Jacobians. \\
    $\boldsymbol{\tau}$  & Joint torque vector. \\
    $\boldsymbol{\tau}_e$  & Equilibrating joint torque vector. \\
    $\boldsymbol{\tau}_m$  & Manipulating joint torque vector. \\
    $\mb{h}_o$ & Resultant wrench applied by the manipulator to its environment. Includes resultant force $\mb{f}_o$ and resultant torque $\mb{t}_o$. \\
    $\mb{h}_i$  & Wrench applied to the end-effector by the $i$-th leg. Includes a pure force $\mb{f}_i$ and a pure torque $\mb{t}_i$ \\
    $\mb{h}$  & Stacked vector of applied wrenches $\mb{h} = \begin{bmatrix} \mb{h}_1^T & \ldots & \mb{h}_n^T \end{bmatrix}^T$. \\
    $\mb{h}_m$  & Manipulating wrench distribution. \\
    $\mb{f}_e$  & Equilibrating force distribution. \\
    $\mb{G}$  & Grasp matrix. \\
    $\mb{B}$  & Vector basis matrix. \\
    $\mb{M}_o$  & Mass matrix of the end-effector. \\
    $\mb{M}$  & Mass matrix of the system of Lumped Mass-Inertia Elements (LMIEs). \\
    $m_i^*$  & Virtual mass of the $i$-th LMIE. \\
    $\mb{I}_i^*$  & Virtual inertia tensor of the $i$-th LMIE. \\
    $\mb{1}_{\_}$  & Identity matrix where an added subscript will specify the dimension. \vspace{0.25em} \\
    \hline
\end{tabular}

Modern robotics relies increasingly on redundant actuation ---where the number of actuators exceeds the dimension of the task space---to circumvent the degeneration of the kinematic and static equations and improve performance. Redundancy can be leveraged to improve grip tightness in grasping, optimize energy consumption in walking robots, or ensure the proper cooperation of cooperating manipulators. Similarly, in parallel manipulators, it allows for singularity avoidance, homogenized load distributions, and increased feasible wrench capability \cite{Gossel2018}.

In systems governed by the grasp matrix, redundant actuation causes the linear system in (\ref{eq:GraspStatics}) to become indeterminate. For a given input vector $\mb{h}$, the resultant wrench $\mb{h}_o$ is uniquely determined, but there exist infinitely many solutions for $\mb{h}$ that can generate it. The components of $\mb{h}$ that lie in the null-space of $\mb{G}$ have no effect on the resultant wrench but contribute to the \say{squeezing} of the object \cite{Walker1991}. 

In parallel manipulators, redundancy is more nuanced. It may appear as kinematic redundancy (extra degrees of freedom allowing internal self-motion), sensing redundancy (additional sensors for control), or actuation redundancy \cite{Gossel2018}. For the purpose of the analyses in this paper, we are strictly concerned with actuation redundancy since this is the only type that overconstrains the end-effector. Much like the grasp case, this renders the parallel manipulator statically indeterminate: components in the null-space of the Jacobian matrices contribute to the squeezing of the end-effector without inducing motion.

Historically, the analysis of these null-space wrench components was established for grasp-like systems and only later adapted to parallel manipulators. However, we demonstrate that these adaptations overlook fundamental differences in the wrench mappings of these types of robots. Existing approaches often naively study the Jacobian-based wrench mappings of parallel manipulators through grasp matrix frameworks, leading to physically erroneous conclusions regarding internal loading. 

In this paper, we aim to highlight and correct these oversights. We identify the shortcomings of existing approaches and propose a generalized framework that properly extends methods developed for grasp matrix applications to parallel manipulators with actuation redundancy. Our generalized framework ensures that the distinct nature of the wrench mapping defined by the Jacobian matrices is preserved and returns results which are consistent with existing definitions of null-space wrench components.

\section{Related Work}
\label{sec:RelatedWork}
Null-space wrench components have been extensively studied in the context of redundantly-actuated robotic systems. Early work by Kumar and Waldron \cite{Kumar1988} established a geometric foundation for the analysis of multifingered grippers and walking robots. They defined the \emph{interaction force} between two forces as the difference in their components along the line joining their application points. Mathematically, a system is free of interaction forces if, for every pair of applied forces $\mb{f}_i$ and $\mb{f}_j$ at locations $\mb{r}_i$ and $\mb{r}_j$:

\begin{equation}
    (\mb{f}_j - \mb{f}_i) \cdot (\mb{r}_j - \mb{r}_i) = 0.
    \label{eq:KWInteractionForce}
\end{equation}
They termed the set of forces satisfying this condition the \emph{equilibrating force distribution}. To eliminate the components in the null-space of the grasp matrix, which are known to cause internal stress, and determine the equilibrating force distribution $\mb{f}_e$, they proposed the unweighted Moore-Penrose pseudo-inverse solution 

\begin{equation}
    \mb{f}_e = \mb{G}^{\dagger} \mb{h}_o
    \label{eq:KWMinNormSolution}
\end{equation}
where $\mb{f}_e = \begin{bmatrix}
    \mb{f}_{e,1}^T & \ldots & \mb{f}_{e,n}^T
\end{bmatrix}^T$ is the stacked vector of equilibrating forces and $\mb{G}^{\dagger} = \mb{G}^T (\mb{G} \mb{G}^T)^{-1}$.

Subsequent research sought to replace geometric definitions with physically meaningful interpretations of null-space wrench components. This often involved constructing virtual physical models to represent stress within the handled object. Notable examples include the \emph{virtual sticks} formulation proposed by Uchiyama and Dauchez \cite{Uchiyama1988} and the \emph{virtual linkage model} proposed by Williams and Khatib \cite{Williams1993}. During this time, null-space wrench components were more often referred to as \emph{internal loads}. Some authors correctly distinguished internal loads from interaction forces, though the terms were often conflated. 

Walker et al. were among the first to highlight the distinction between interaction forces and internal loads in \cite{Walker1991}. In their analysis of cooperating serial manipulators, they introduced a \say{nonsqueezing} generalized inverse of the grasp matrix, claiming it yielded wrench distributions free of internal loads. However, the validity of their approach was later contested. Chung et al. \cite{Chung2005} demonstrated that the method proposed in \cite{Walker1991} did, in fact, generate internal loads, and instead opted to use the Moore-Penrose pseudo-inverse. Zuo and Qian \cite{Zuo1999} also investigated interaction forces and internal loads by studying the conditions under which these two sets become identical in grasping applications. 

More recently, Erhart and Hirche \cite{Erhart2015} resolved many of these ambiguities by grounding the definition of internal loads in the principles of analytical dynamics, specifically using the Udwadia-Kalaba equation \cite{Udwadia2010}. They demonstrated that the \emph{manipulating wrench distribution}---the set of wrenches generating a resultant wrench $\mb{h}_o$ without generating internal loads---has infinitely many solutions in some cases. To address this, they derive a parametrized Moore-Penrose pseudo-inverse of the grasp matrix, $\mb{G}^+_M$, to determine the full solution space. 

While Erhart and Hirche provided a robust theoretical basis, gaps remained regarding the precise characterization of internal loads in general redundant systems. In recent work \cite{Flight2026}, we addressed these limitations by refining the theorem proposed in \cite{Erhart2015} to account for factors that had been overlooked. We also further clarified the distinction between the geometric interaction forces as defined by Kumar and Waldron and constraint wrenches as defined by Udwadia and Kalaba. This revised framework allowed us to fully characterize the uniqueness of the manipulating wrench distribution, refute the primary assumption in the leading wrench decomposition algorithms, and derive explicit, general solutions to the wrench synthesis and wrench analysis problems. 

Although the theoretical foundations of internal loading are well established for grasp-type systems, a critical gap remains for the applications of these principles to parallel manipulators where the wrench mapping is fundamentally different. 

The wrench capabilities of parallel manipulators have been studied extensively \cite{Krut2004, Nokleby2005, Zibil2007, Firmani2007, Firmani2008a, Firmani2008b, Mejia2015, Mejia2016, Mejia2021}. Some authors proposed optimization-based strategies to determine the feasible wrench sets of a given manipulator \cite{Nokleby2005, Firmani2007, Mejia2015, Mejia2021}. However, as the computational cost of these approaches became apparent, authors gravitated towards explicit analytical methods \cite{Zibil2007, Firmani2008a, Firmani2008b, Mejia2016}. 

For parallel manipulators with actuation redundancy, the central challenge lies in exploiting the additional actuation to increase the wrench capabilities of a manipulator, or to optimize secondary performance objectives with informed use of the null-space wrench components. It is well documented that the maximum output wrench is generally achieved when all actuators operate at their torque limits, even though this inevitably induces substantial null-space wrenches \cite{Nokleby2005, Zibil2007, Firmani2008b}. The physical significance of these components in parallel manipulators, however, is still not well understood. The \emph{scaling factor method} proposed in \cite{Nokleby2005} and used again in \cite{Zibil2007, Garg2009} invokes Kumar and Waldron's method directly, asserting that the Moore-Penrose pseudo-inverse solution for the joint torque vector $\boldsymbol{\tau}$ (i.e., the minimum-norm solution for $\boldsymbol{\tau}$) does not generate internal loads at the end-effector. 

We argue that this approach suffers from two critical oversights. First, it fails to differentiate between interaction forces and internal loads. As originally discussed in grasp matrix research \cite{Walker1991, Zuo1999} and recently clarified in our own work \cite{Flight2026}, these variants of null-space wrench components refer to different physical phenomena. Kumar and Waldron clearly state in their work that their method relates to interaction forces, not internal loads. Second, and more importantly, this adaptation ignores the fundamental differences between the wrench mappings of grasp-like systems and parallel robots. The original method for eliminating interaction forces was established for systems in which wrench equilibrium is defined by the grasp matrix (\ref{eq:GraspStatics}), whereas the wrench equilibrium equation of parallel manipulators invokes the Jacobian matrices of the manipulator. 

Other authors have investigated the disparity between these two kinematic mappings in previous work. A parallel robot was analyzed in \cite{Erden2007} where the authors correctly distinguish between the minimum-norm solution of the joint torque vector $\boldsymbol{\tau}$ and that of the applied force vector $\mb{f}$. However, they do not describe the difference between these two solutions, nor why the minimum-norm solution for $\boldsymbol{\tau}$ generates interaction forces. In \cite{Xu2012}, the authors identify the conditions under which the minimum-norm solution for $\boldsymbol{\tau}$ does not generate internal forces by considering the elastic deformation of the manipulator caused by each constraint and actuation wrench. Although their analysis is sound, we believe their conclusions offer an incomplete description of interaction forces in parallel manipulators and that their results are specific to the manipulator architecture which they consider. 

To the best of our knowledge, the question of how to adapt methods for synthesizing wrench distributions free of interaction forces or internal loads in grasp-like systems to parallel manipulators with actuation redundancy has yet to be answered. The current state of the art for eliminating interaction forces fails to account for the differences in the wrench mappings of the two classes of systems, and a method for calculating valid solutions at the joint torque level has yet to be derived. In addition, we have not found any examples of methods such as the ones proposed in \cite{Erhart2015, Flight2026} for eliminating internal loads being adapted to parallel manipulators. 

In this paper, we aim to address these gaps by extending recent theoretical developments regarding null-space wrench components to redundant parallel manipulators. Building on the results presented in \cite{Flight2026}, we generalize these methods to accommodate the Jacobian-based wrench mapping of parallel robots. We highlight specific instances in the literature where the direct application of grasp matrix methods has led to incorrect conclusions, and we provide corrected results using our generalized methodology.

\section{Theoretical Background}
\subsection{Kinematics of Parallel Manipulators}
Parallel manipulators, whether non-redundant or redundant, comprise multiple closed-loop kinematic chains. The kinematic model for this type of manipulator is often more complex than that of serial manipulators and is therefore expressed as a set of implicit nonlinear equations of the form

\begin{equation}
    \mb{g} ( \boldsymbol{\theta}, \mb{x}_o) = \mb{0}
\end{equation}
where $\boldsymbol{\theta}$ is the $q$-dimensional vector of joint coordinates and $\mb{x}_o$ is the $p$-dimensional vector of Cartesian coordinates. Taking the first time-derivative, we obtain the velocity equations of the manipulator written as 

\begin{equation}
    \mb{J} \dot{\mb{x}}_o = \mb{K} \dot{\boldsymbol{\theta}}
    \label{eq:parRobotKinematics}
\end{equation}
where $\mb{J}$ and $\mb{K}$ are the Jacobian matrices, $\dot{\mb{x}}$ is the Cartesian velocity vector and $\dot{\boldsymbol{\theta}}$ is the joint velocity vector. For parallel manipulators with actuation redundancy, matrix $\mb{J}$ is typically $q \times p$ and matrix $\mb{K}$ is typically $q \times q$.

\subsection{Wrench Transmission in Parallel Manipulators}
The wrench equilibrium equation for parallel manipulators can be obtained with the static-kinematic duality. Assuming quasi-static conditions, the power balance of the manipulator can be written as 

\begin{equation}
    \mb{h}_o^T \dot{\mb{x}}_o = \boldsymbol{\tau}^T \dot{\boldsymbol{\theta}}
    \label{eq:powerBalance}
\end{equation}
where $\mb{h}_o = \begin{bmatrix}
    \mb{f}_o^T & \mb{t}_o^T
\end{bmatrix}^T$ is the resultant wrench applied to the end-effector defined as in (\ref{eq:GraspStatics}), and $\boldsymbol{\tau}$ is the joint torque vector. Substituting (\ref{eq:parRobotKinematics}) into (\ref{eq:powerBalance}), we obtain

\begin{equation}
    \mb{h}_o^T \dot{\mb{x}}_o = \boldsymbol{\tau}^T \mb{K}^{-1} \mb{J} \dot{\mb{x}}_o.
\end{equation}
For this equation to hold for all $\dot{\mb{x}}_o$, we must have

\begin{equation}
    \mb{h}_o = \mb{J}^T \mb{K}^{-T} \boldsymbol{\tau},
    \label{eq:parRobotStatics}
\end{equation}
which maps the joint torques $\boldsymbol{\tau}$ to the resultant wrench $\mb{h}_o$ and is the wrench equilibrium equation for parallel manipulators. 

It is important for what will follow to understand the physical significance of this equation. In fact, (\ref{eq:parRobotStatics}) reveals the two-stage wrench transmission scheme of parallel manipulators. Each actuated joint adds one constraint to the end-effector, and matrix $\mb{J}^T$ defines the \emph{external geometry} of these constraints, where the $i$-th column represents the screw axis of the wrench applied by the $i$-th actuated joint to the end-effector. 

Meanwhile, $\mb{K}^{-T}$ defines the \emph{internal mechanical advantage} of the actuators. It transforms the input joint efforts in $\boldsymbol{\tau}$ into scalar coefficients that scale the columns of $\mb{J}^T$. These scalar coefficients act as transmission weights which determine to what extent a force or torque applied by an actuator is felt at the end-effector. Physically, matrix $\mb{K}^{-T}$ determines how effectively a given actuator can generate a force along its corresponding constraint line. 

\subsection{Standard Redundancy Resolution}
For a $p$-dof parallel manipulator with $q$ actuators, actuation redundancy is attained when $q > p$ \cite{Gossel2018}. For these types of manipulators, the linear system defined by (\ref{eq:parRobotStatics}) becomes indeterminate, and infinitely many joint effort vectors $\boldsymbol{\tau}$ can produce the same resultant wrench $\mb{h}_o$. The standard redundancy resolution method is to use an equation of the form 

\begin{equation}
    \boldsymbol{\tau} = (\mb{J}^T \mb{K}^{-T})^+ \mb{h}_o + [\mb{1}_q - (\mb{J}^T \mb{K}^{-T})^+(\mb{J}^T \mb{K}^{-T})] \mb{z}
\end{equation}
where $(\mb{J}^T \mb{K}^{-T})^+$ is any generalized inverse of the linear mapping $\mb{J}^T \mb{K}^{-T}$, $\mb{1}_q$ is the $q \times q$ identity matrix, and $\mb{z}$ is an $q$-dimensional vector which defines the location of a given solution in the null space of $\mb{J}^T \mb{K}^{-T}$. The components in the null space have no effect on the resultant wrench, but can be used to optimize secondary criteria such as singularity avoidance or load balancing among the actuators.

\section{Definition of Interaction Forces and Internal Loads}
\label{sec:InteractionForcesInternalLoadsParManips}
Evidently, redundant actuation and the possibility of generating null-space wrench components adds an additional layer of complexity to the control of redundant robots. This has motivated many researchers to study the physical implications of these null-space components and to propose methods for wrench synthesis that aim to limit or exploit them. However, such methods are typically proposed for grasp-like systems with wrench equilibrium equations of the form shown in (\ref{eq:GraspStatics}). 

Though their fundamental equations may differ, parallel robots are analogous to grasp-like systems, since we may consider the end-effector as being \say{grasped} by the legs of the manipulator. Here, we give the form of the grasp matrix and define the two leading characterizations of null-space wrench components before discussing their application to parallel manipulators.


\subsection{The Grasp Matrix}
Arbitrary wrenches applied to a rigid body and their internal and external components can be studied using the grasp matrix $\mb{G}$ which maps the applied wrenches to the resultant wrench that they generate as shown in (\ref{eq:GraspStatics}). The form of $\mb{G}$ depends on the contact model. 

\paragraph*{Case 1 - Pure Forces (Point Contact)} 
\textcolor{black}{The kinematic chains only apply pure forces. Here, $\mb{h} = \begin{bmatrix}
    \mb{f}_1^T & \ldots & \mb{f}_n^T
\end{bmatrix}^T \in \mathbb{R}^{3n}$ and $\mb{G}$ takes the form} 

\begin{equation}
    \mb{G} = \begin{bmatrix}
        \mb{1}_3 & \cdots & \mb{1}_3 \\
        \mb{S}(\mb{r}_1) & \cdots & \mb{S}(\mb{r}_n)
    \end{bmatrix}
    \label{eq:GPureForces}
\end{equation}

\paragraph*{Case 2 - General Wrenches (Rigid Contact)} 
\textcolor{black}{The kinematic chains apply both forces and torques. Here, $\mb{h} = \begin{bmatrix}
    \mb{f}_1^T & \mb{t}_1^T & \ldots & \mb{f}_n^T & \mb{t}_n^T
\end{bmatrix}^T \in \mathbb{R}^{6n}$ and $\mb{G}$ becomes}

\begin{equation}
    \mb{G} = \begin{bmatrix}
        \mb{1}_3 & \mb{0}_3 & \cdots & \mb{1}_3 & \mb{0}_3 \\ 
        \mb{S}(\mb{r}_1) & \mb{1}_3 & \cdots & \mb{S}(\mb{r}_n) & \mb{1}_3
    \end{bmatrix}.
    \label{eq:GForcesandMoments}
\end{equation} 

Note that $\mb{S}(\mb{r}_i)$ denotes the skew-symmetric matrix that represents the cross product with vector $\mb{r}_i$ such that $\mb{S}(\mb{r}_i) \mb{a} = \mb{r}_i \times \mb{a}$ where $\mb{a}$ can be any 3-vector and that $\mb{S}(\mb{r}_i)^T = -\mb{S}(\mb{r}_i)$ is a property of skew-symmetric matrices.







\subsection{Definition of Interaction Forces}
Interaction forces were first formally defined by Kumar and Waldron in \cite{Kumar1988}. The authors consider only pure forces (no torques applied at the contact point) and provide the geometric condition shown in (\ref{eq:KWInteractionForce}).

It has since been noted by some authors that there is no clear way to extend the analogy of interaction forces to pure torques \cite{Erhart2015}. Therefore, to remain within the intended domain of applicability of Kumar and Waldron's method, we restrict our consideration of interaction forces to situations in which only pure forces are applied to the rigid body. 

In \cite{Kumar1988}, it was shown that the equilibrating force distribution is the solution which minimizes the Euclidean norm of the stacked vector of applied forces $||\mb{f}||_2$. This is a typical quadratic optimization problem with equality constraints. We can write this as

\begin{equation}
    \begin{aligned}
        \min & \qquad Z =  \frac{1}{2}\mathbf{f}^T \mathbf{f}\\ 
        \text{subject to}& \qquad \mathbf{h}_o = \mathbf{G} \mathbf{f}.
        \label{eq:InteractionForceOptProb}
    \end{aligned}
\end{equation}

The equilibrating force distribution is the vector $\mb{f}$ which solves (\ref{eq:InteractionForceOptProb}) and can be obtained explicitly with (\ref{eq:KWMinNormSolution}).

\subsection{Definition of Internal Loads}
\label{sec:Def-InternalLoads}
We recently clarified the distinction between interaction forces and internal loads and proposed explicit solutions to the wrench synthesis and wrench analysis problems in \cite{Flight2026}. The definition of internal loads is derived from the Udwadia-Kalaba equation \cite{Udwadia2010} which can be written as

\begin{equation}
    \mb{h}_m = \mb{h} + \mb{h}_c
\end{equation}
where $\mb{h}_m = \begin{bmatrix}
    \mb{h}_{m,1} & \ldots & \mb{h}_{m,n}
\end{bmatrix}$ is the stacked vector of manipulating wrenches and $\mb{h}_c = \begin{bmatrix}
    \mb{h}_{c,1} & \ldots & \mb{h}_{c,n}
\end{bmatrix}$ is the stacked vector of constraint wrenches. 

This equation is essentially a reformulation of Gauss's principle of least constraint. If a resultant wrench $\mb{h}_o$ is applied to the CoM of a rigid body of mass $m_o$ and inertia tensor $\mb{I}_o$, it will undergo an instantaneous acceleration given by

\begin{equation}
\label{eq:RBDynamics}
    \ddot{\mb{x}}_o = \begin{bmatrix}
        \ddot{\mb{p}}_o \\ \dot{\boldsymbol{\omega}}_o
    \end{bmatrix} = \mb{M}_o^{-1} \mb{h}_o, \quad \mb{M}_o = \begin{bmatrix}
        m_o \mb{1}_3 & \mb{0} \\
        \mb{0} & \mb{I}_o
    \end{bmatrix}
\end{equation}
where $\ddot{\mb{x}}_o = \begin{bmatrix}
    \ddot{\mb{p}}_o^T & \dot{\boldsymbol{\omega}}_o^T
\end{bmatrix}^T$ is the instantaneous acceleration induced by $\mb{h}_o$ with $\ddot{\mb{p}}_o$ and $\dot{\boldsymbol{\omega}}_o$ being the acceleration and angular acceleration, respectively. $\mb{M}_o$ is the generalized inertia matrix about the CoM of the rigid body.

\begin{figure}
    \centering    \includegraphics[width=0.5\linewidth]{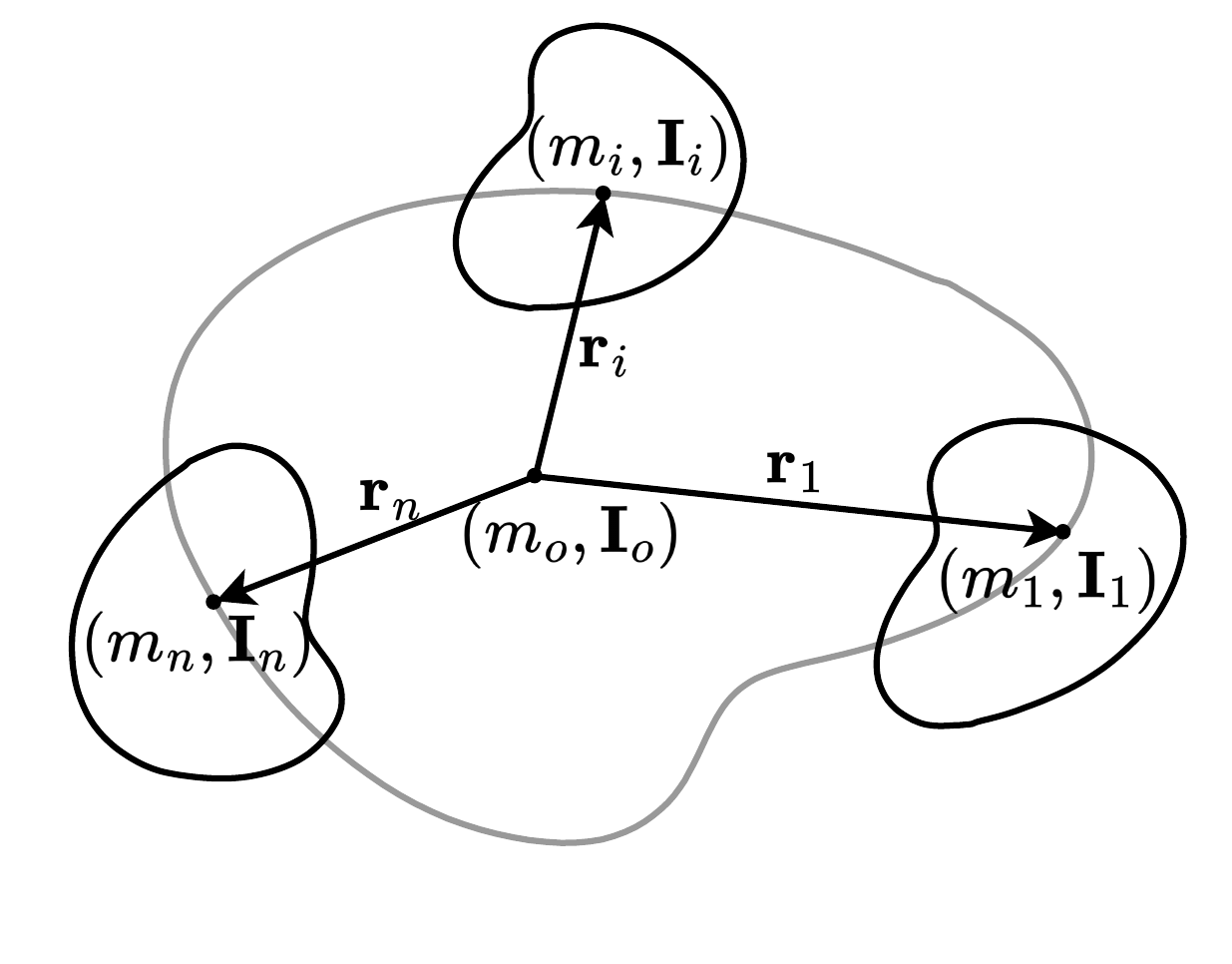}
    \caption{Example of a dynamically equivalent system of $n$ LMIEs.}
    \label{fig:equimomentalSystem}
\end{figure}

To calculate a set of wrenches that generates a resultant wrench while eliminating or prescribing internal loads, or to analyze a known set of applied wrenches, we must first determine a valid dynamically equivalent system for the rigid body. A dynamically equivalent system is a collection of $n$ rigidly connected discrete elements with mass $m_i$ and inertia tensor $\mb{I}_i$ that, together, have the same mass, centre of mass, and inertia tensor as the physical object (see Fig. \ref{fig:equimomentalSystem}). As in \cite{Flight2026}, we call these discrete elements Lumped Mass-Inertia Elements (LMIEs) and define them as rigid bodies of arbitrary shape with known mass $m_i$ and inertia tensor $\mb{I}_i$. A dynamically equivalent system of $n$ LMIEs each with properties $(m_i, \mb{I}_i)$ for a rigid object with properties $(m_o, \mb{I}_o)$ must satisfy

\begin{align}
    m_o &= \sum_{i=1}^n m_i \  \text{(mass equivalence)}, \label{eq:MassSumEquivalence}\\
    \mb{I}_o &= \sum_{i=1}^n \mb{I}_i + \sum_{i=1}^n \mb{S}(\mb{r}_i) m_i \mb{S}(\mb{r}_i)^T \ \text{(inertia equiv.)}, \label{eq:MomentofInertiaEquivalence}\\
    \sum_{i=1}^n & \mb{r}_i m_i = \mb{0}_{3 \times 1} \ \text{(CoM equivalence)}. \label{eq:CoMEquivalence}
\end{align}

The wrench $\mb{h}_i$ applied at $\mb{r}_i$ will induce an unconstrained acceleration of the corresponding LMIE given by

\begin{equation}
    \label{eq:LMIEDynamics}
        \ddot{\mb{x}}^d_i = \mb{M}_i^{-1} \mb{h}_i, \quad \mb{M}_i = \begin{bmatrix}
            m_i \mb{1}_3 & \mb{0} \\
            \mb{0} & \mb{I}_i
        \end{bmatrix}
    \end{equation}
where $\ddot{\mb{x}}^d_i$ and $\mb{M}_i$ are the unconstrained acceleration and inertia matrix of the $i$-th LMIE, respectively. Equation (\ref{eq:LMIEDynamics}) returns the acceleration that an LMIE \emph{would have had} subject to applied wrench $\mb{h}_i$, had it not been fixed to the rigid body. 

\begin{figure}
    \centering    \includegraphics[width=0.65\linewidth]{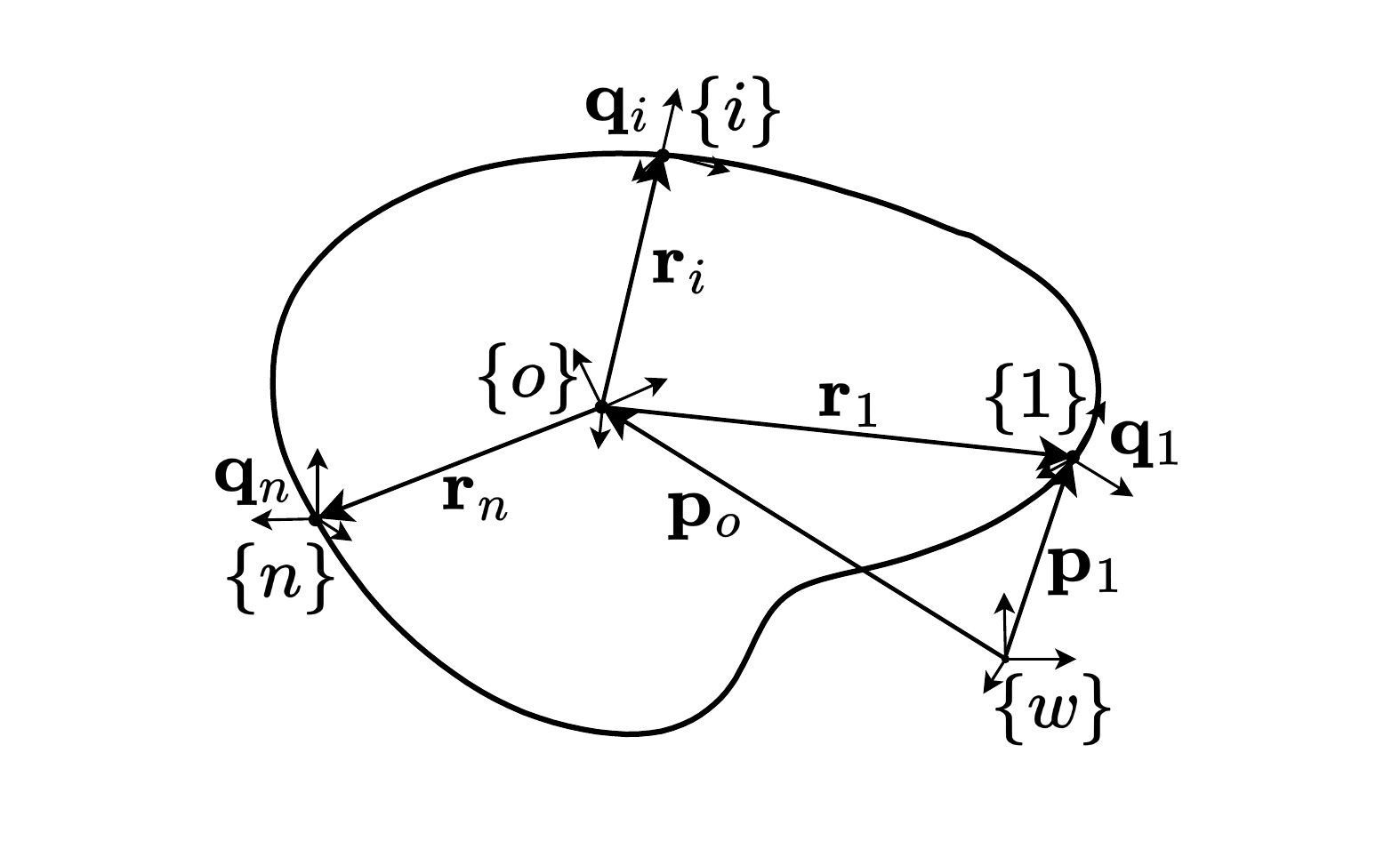}
    \caption{Kinematic quantities.}
    \label{fig:KinematicQuantities}
\end{figure}

We use vectors $\mb{p}_i$ and $\mb{q}_i$ to define the position and orientation of the $i$-th LMIEs reference frame. Since all LMIEs lie on the same rigid body, we consider these reference frames to be fixed to the rigid body and the relative position and orientation of the LMIE frames must remain constant (see Fig. \ref{fig:KinematicQuantities} for the relevant kinematic quantities). We can write these kinematic constraints in matrix form as

\begin{equation}
    \mb{A} \ddot{\mb{x}} = \mb{b},
    \label{eq:KinConstraints}
\end{equation}
where $\ddot{\mb{x}} = \begin{bmatrix}
    \ddot{\mb{x}}_1^T & \ldots & \ddot{\mb{x}}_n^T
\end{bmatrix}^T$ is the stacked vector of constrained LMIE accelerations where each $\ddot{\mb{x}}_i$ is defined similarly to $\ddot{\mb{x}}_o$, and

\begin{equation}
    \mb{A} = \begin{bmatrix}
        -\mb{1}_3 & \mb{S}(\mb{r}_{21}) & \mb{1}_3 & \mb{0} &  &  & \\
        \mb{0} & -\mb{1}_3 & \mb{0} & \mb{1}_3 &  &  & \\
        \vdots & \vdots &  &  & \ddots & & \\
        -\mb{1}_3 & \mb{S}(\mb{r}_{n1}) &  &  &  & \mb{1}_3 & \mb{0} \\
        \mb{0} & -\mb{1}_3 &  &  &  & \mb{0} & \mb{1}_3
    \end{bmatrix}
\end{equation}
where $\mb{S}(\cdot)$ is a skew-symmetric matrix representing a cross product with the argument. $\mb{r}_{ji}$ is a relative position vector from $\mb{r}_i$ to $\mb{r}_j$. The centripetal terms are grouped in vector $\mb{b}$ which is written as

\begin{equation}
    \mb{b} = \begin{bmatrix}
        \mb{S}(\boldsymbol{\omega}_2)^2 \mb{r}_{21} \\ \mb{0} \\ \vdots \\ \mb{S}(\boldsymbol{\omega}_n)^2 \mb{r}_{n1} \\ \mb{0}
    \end{bmatrix}.
\end{equation}

\begin{figure}
    \subfloat[Three wrenches $\mb{h}_i$ applied to a rigid body generate a resultant wrench $\mb{h}_o$.]{%
        \includegraphics[width=.3\linewidth]{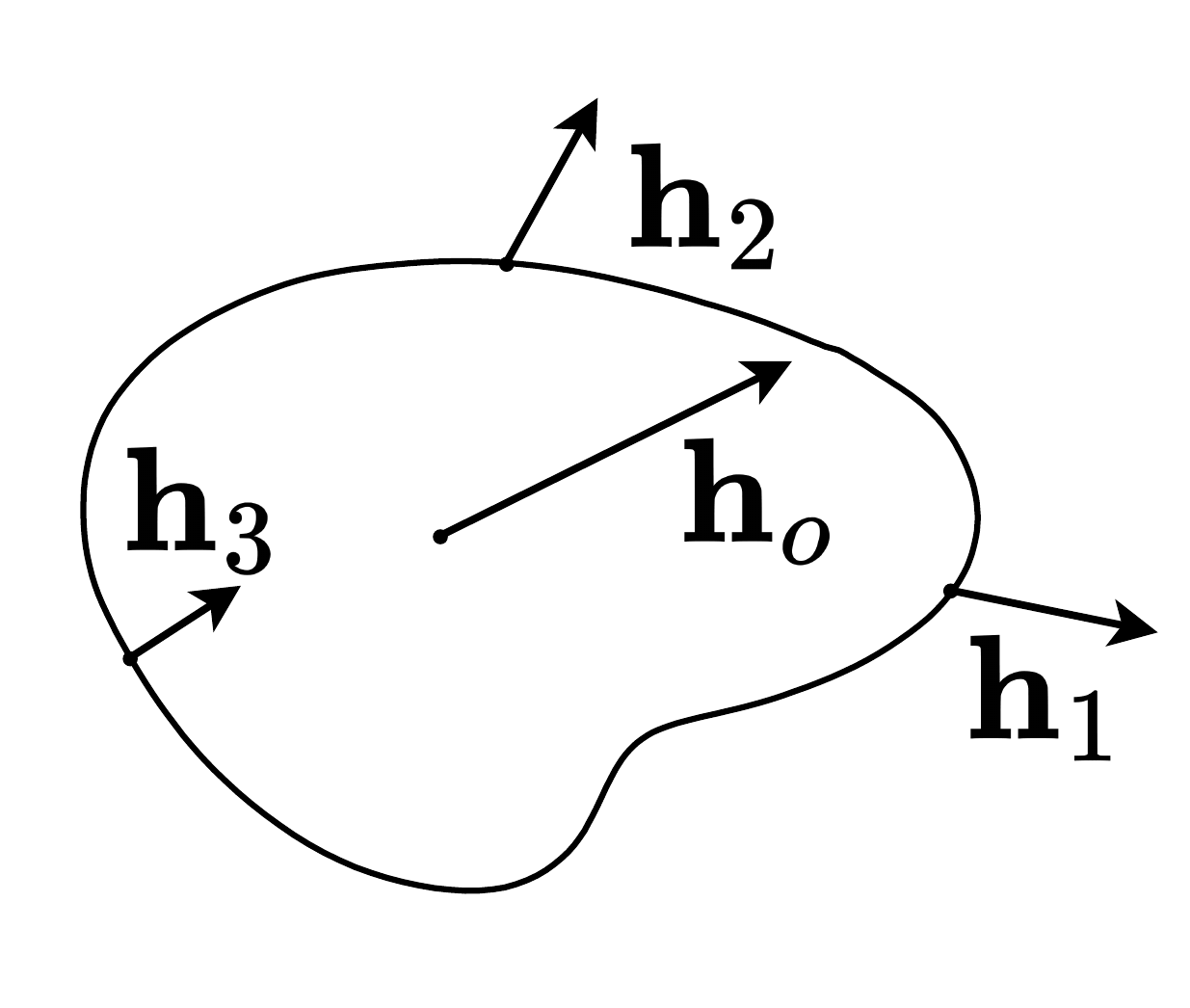}%
        \label{subfig:ConstrainedAccelerations1}%
    } \
    \subfloat[The resultant wrench $\mb{h}_o$ induces rigid body acceleration $\ddot{\mb{x}}_o$.]{%
        \includegraphics[width=.3\linewidth]{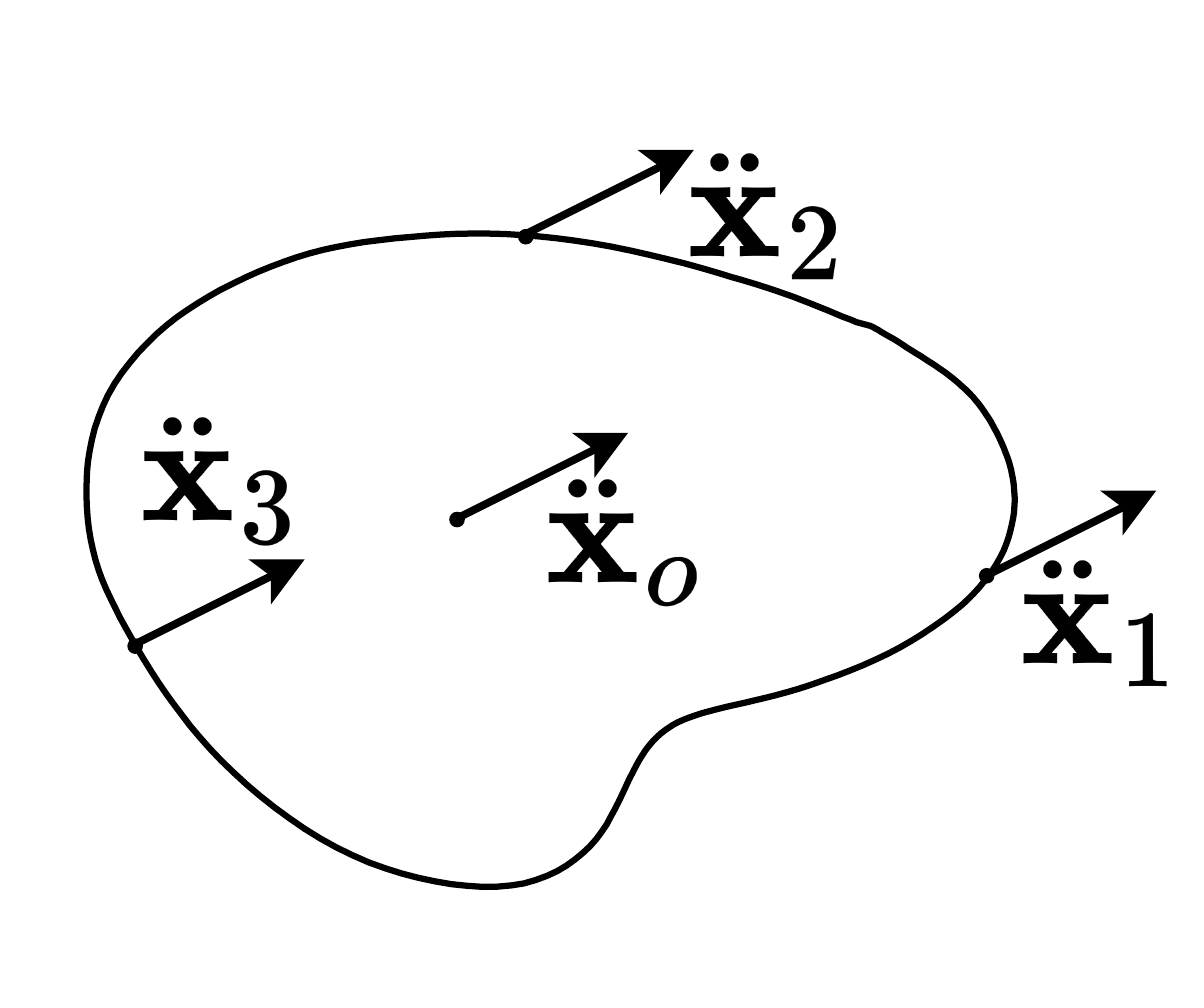}%
        \label{subfig:ConstrainedAccelerations2}%
    }\
    \centering
    \subfloat[Constraint wrenches $\mb{h}_{c,i}$ ensure compliance with rigid body motion.]{%
        \includegraphics[width=.33\linewidth]{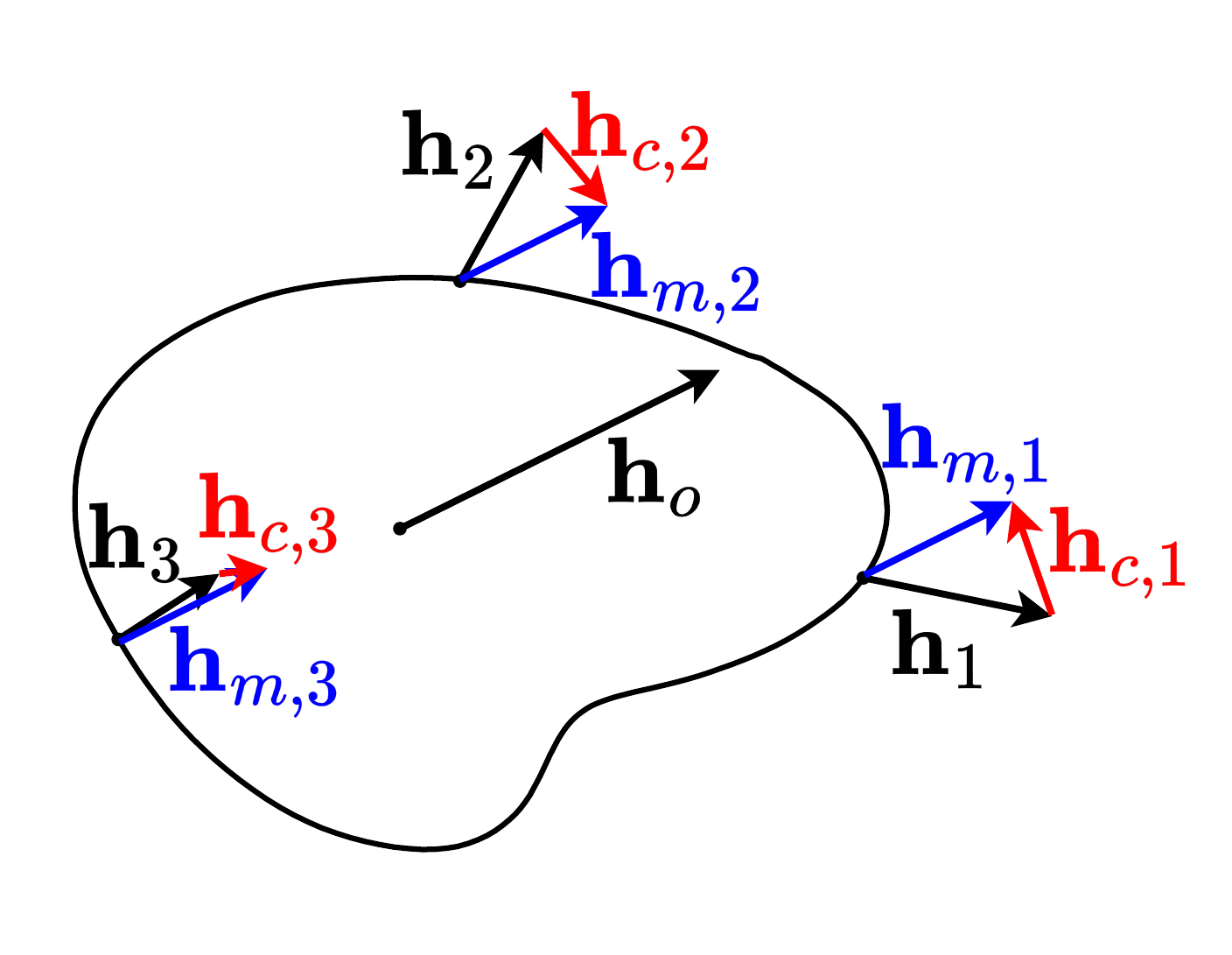}%
        \label{subfig:ConstrainedAccelerations3}%
    }
    \caption{Illustration of physical the significance of constraint wrenches.}
    \label{fig:ConstrainedAccelerations}
\end{figure}

When the unconstrained acceleration $\ddot{\mb{x}}^d_i$ of an LMIE induced by $\mb{h}_i$ violates the kinematic constraints (i.e., is not consistent with the constrained motion of the rigid body), a constraint wrench $\mb{h}_{c, i}$ is added such that the resultant is the manipulating wrench $\mb{h}_{m,i}$ needed to generate the appropriate constrained acceleration that satisfies (\ref{eq:KinConstraints}). This is illustrated in Fig. \ref{fig:ConstrainedAccelerations}.

Like the equilibrating force distribution, the manipulating wrench distribution can also be written as an optimization problem. In this case, it is defined as

\begin{equation}
    \begin{aligned}
        \min& \qquad Z = \frac{1}{2} \mb{h}^T \mb{M}^{-1} \mb{h}\\ 
        \text{subject to}& \qquad \mb{h}_o = \mb{G} \mb{h}.
        \label{eq:InternalLoadOptProb}
    \end{aligned}
\end{equation}
where $\mb{M}$ is the inertia matrix of the system. $\mb{M}$ is a diagonal block matrix written as

\begin{equation}
    \mb{M} = \begin{bmatrix}
        m_1 \mb{1}_3 & \mb{0} & \cdots & \mb{0} & \mb{0} \\
        \mb{0} & \mb{I}_1 & \cdots & \mb{0} & \mb{0} \\
        \vdots & \vdots & \ddots & \vdots & \vdots \\
        \mb{0} & \mb{0} & \cdots & m_n \mb{1}_3 & \mb{0} \\
        \mb{0} & \mb{0} & \cdots & \mb{0} & \mb{I}_n
    \end{bmatrix}.
\end{equation}
The vector $\mb{h}$ which solves (\ref{eq:InternalLoadOptProb}) is obtained with

\begin{equation}
    \mb{h}_m = \mb{G}^+_M \mb{h}_o
\end{equation}
where $\mb{G}^+_M$ is the parametrized Moore-Penrose pseudo-inverse of the grasp matrix written as 

\begin{equation}
    \mb{G}^+_M = \begin{bmatrix}
        \frac{m^*_1}{m^*_o} \mb{1}_3 & m^*_1 \mb{S}(\mb{r}_1)^T [\mb{I}^*_o]^{-1}  \\
        \mb{0}_3 & \mb{I}^*_1 [\mb{I}^*_o]^{-1} \\
        \vdots & \vdots \\
        \frac{m^*_n}{m^*_o} \mb{1}_3 & m^*_n \mb{S}(\mb{r}_n)^T [\mb{I}^*_o]^{-1}  \\
        \mb{0}_3 & \mb{I}^*_n [\mb{I}^*_o]^{-1}
    \end{bmatrix}.
    \label{eq:ParametrizedInverse}
\end{equation}

As shown in \cite{Flight2026}, the virtual inertia parameters $(m^*_i, \mb{I}^*_i)$ used to span the solution space for the manipulating wrench distribution must satisfy

\begin{align}
    m^*_o &= \sum_{i=1}^n m^*_i, \label{eq:virtualMassSum}\\
    \mb{I}^*_o &= \sum_{i=1}^n \mb{I}_i^* + \sum_{i=1}^n \mb{S}(\mb{r}_i) m^*_i \mb{S}(\mb{r}_i)^T, \label{eq:inertiaTensorEquivalence}\\
    \quad & \sum_{i=1}^n \mb{r}_i m^*_i = \mb{0}, \label{eq:virtualCoMEquivalence}\\
    \mb{I}^*_i \ \propto& \sum_{i=1}^n \mb{S}(\mb{r}_i) m^*_i \mb{S}(\mb{r}_i)^T \propto \ \mb{I}^*_o, \ \forall i. \label{eq:inertiaProportionality}
\end{align}

The results presented in \cite{Flight2026} also had important implications for our understanding of manipulating wrench distributions. These are summarized here as they are important for the case studies discussed in this paper: 

\begin{enumerate}
    \item \textbf{Proportionality of the inertial parameters.} To calculate the manipulating wrench distribution $\mb{h}_m$ for a given resultant wrench $\mb{h}_o$, we must first define a dynamically equivalent system of $n$ LMIEs. This system is considered to be at rest at the time-step under consideration, even if the actual rigid body is moving. An interesting result of this assumption is that the inertial properties of the dynamically equivalent system $(m^*_o, \mb{I}^*_o)$ need only be \emph{proportional} to the actual inertial properties of the end-effector $(m_o, \mb{I}_o)$.
    \item \textbf{Torque weighting.} The manipulating wrench distribution is not unique in many cases, and it is never unique when pure torques can be applied to the end-effector. This is because a variable amount of the resultant torque $\mb{t}_o$ can be assigned to the manipulating torques $\mb{t}_{m,i}$. We are able to control which portion is generated by the manipulating torques by modulating the contribution of the inertia tensors of the LMIEs $\mb{I}^*_i$ to the total inertia of the dynamically equivalent system $\mb{I}^*_o$ compared to that of the system of point masses $\sum_{i=1}^n \mb{S}(\mb{r}_i) m^*_i \mb{S}(\mb{r}_i)^T$. The larger the contribution from the inertia tensors, the larger the portion of $\mb{t}_o$ that will be generated by the manipulating torques. 
    \item \textbf{Parallel manipulating torques.} For a set of applied torques to belong to the manipulating wrench distribution, they must be parallel to $\mb{t}_o$. This includes the manipulating torques $\mb{t}_{m,i}$ and the manipulating force-induced torque $\mb{t}_{fm}$ which is the total torque generated by the pure forces applied to the end-effector.
    \item \textbf{Uniqueness.} The uniqueness of the manipulating wrench distribution was completely characterized in \cite{Flight2026}. It was found that if the attachment points to the end-effector are pre-defined, a total of 10 pure forces must be applied to a rigid body to obtain a truly valid manipulating wrench distribution in the most general case. This is rarely possible in practice and heavily constrains the solution space. However, by eliminating the most restrictive constraint (\ref{eq:inertiaTensorEquivalence}), we are able to solve a reduced problem as long as the center of mass of the end-effector lies within the convex hull formed by the attachment points of the legs to the end-effector. The problem can be solved with only three applied forces with this simplification, but does generate a small amount of internal loads.
\end{enumerate}

For further details on the results summarized above, the reader is referred to \cite{Flight2026}.

\section{Requirements for Managing Interaction Forces and Internal Loads in Parallel Manipulators}
\label{sec:jointTorques2EEForces}
\subsection{Static Determinism of the Legs}
It was shown in Section \ref{sec:InteractionForcesInternalLoadsParManips} that the definitions of interaction forces and internal loads have physical meaning as they correspond to the minimization of the unweighted and weighted norms of the wrenches applied to the rigid body, respectively. The solutions define the force and/or torque vectors that must be applied to the rigid body by the kinematic chains to eliminate interaction forces or internal loads. However, the typical static mapping (\ref{eq:parRobotStatics}) of parallel manipulators does not reveal the applied wrenches generated by the actuators. Instead, it directly maps the joint torques to the resultant wrench, without consideration for the forces and torques applied by each leg to the end-effector. We therefore require a different mapping for relating the torques generated by the actuated joints on each leg to the wrench components that they apply to the end-effector. For one leg, this is written as

\begin{equation}
    \mb{h}_i = \mb{B}_i \mb{K}_i^{-T} \boldsymbol{\tau}_i
\end{equation}
where $\boldsymbol{\tau}_i$ is a vector containing the torques applied by actuators on the $i$-th leg, $\mb{K}_i^{-T}$ is a sub-matrix of matrix $\mb{K}^{-T}$ that converts the joint torques $\boldsymbol{\tau}_i$ to transmission weights, and the columns of matrix $\mb{B}_i$ form a vector basis for the wrench applied by the $i$-th leg. When we collect these expressions for all the legs of the manipulator, we obtain

\begin{equation}
    \mb{h} = \mb{B} \mb{K}^{-T} \boldsymbol{\tau}.
    \label{eq:jointTorque2AppliedWrench}
\end{equation}

With this equation, it is straightforward to adapt the concepts of interaction forces and internal loads to parallel manipulators and establish the conditions for doing so. We simply obtain the solution for $\mb{h}$ which eliminates interaction forces or internal loads and calculate the corresponding vector of joint torques required to generate these applied wrenches by inverting (\ref{eq:jointTorque2AppliedWrench}). When doing so, three cases may be encountered: 

\paragraph*{Case 1: Statically determined leg.} When a leg has as many actuators as there are components in the wrench $\mb{h}_i$ it applies to the end-effector and its corresponding Jacobian is of full rank, it is \emph{statically determined.} The leg can therefore produce a wrench in any direction in the required wrench space and is able to generate the required equilibrating force or manipulating wrench. In this case, matrix $\mb{B}_i$ is square. Note that matrix $\mb{K}^{-T}$ is square for all the cases listed here, since this is a property of parallel manipulators with actuation redundancy. 

Since matrix $\mb{B}_i$ is square, the mapping between $\boldsymbol{\tau}_i$ and $\mb{h}_i$ is one-to-one, and the vector $\boldsymbol{\tau}_i$ that generates the required wrench $\mb{h}_i$ is uniquely determined. Examples of such a leg would be a leg with 3 actuators that applies a pure force in $\mathbb{R}^3$ to the end-effector, but no torque, or a leg with 6 actuators that applies both a force and a torque both in $\mathbb{R}^3$.

\paragraph*{Case 2: Statically underdetermined leg.} If a leg has fewer actuators than there are components in the wrench it applies to the end-effector, it is \emph{statically underdetermined}. Matrix $\mb{B}_i$ becomes a tall matrix, and the inversion of (\ref{eq:jointTorque2AppliedWrench}) becomes a least-squares problem. In this case, the leg of the manipulator will likely not be able to generate the wrench required to eliminate either interaction forces or internal loads, with the exception of very few specific cases where the required wrench lies in the domain in which the leg \emph{can} produce the required wrench at which point the overdetermined linear system becomes consistent and the least-squares solution returns zero error. 

\paragraph*{Case 3: Statically overdetermined leg.}
The final case involves legs with more actuators than there are components in the wrench it applies to the end-effector. Matrix $\mb{B}_i$ becomes a wide matrix, and the leg is redundant. Similarly to Case 1, the leg will be able to generate the required wrenches. However, additional considerations will have to be taken in order to manage the redundancy of the leg to avoid generating excessive internal stress within the leg. 

\begin{figure}
\centering
    \subfloat[4-\underline{R}RR manipulator with statically underdetermined legs. Inverting (\ref{eq:jointTorque2AppliedWrench}) using the Moore-Penrose pseudoinverse returns a solution which minimizes the error between the desired solution $\mb{f}_e$ shown in blue and the nearest feasible solution $\mb{f}$ shown in red.]{%
        \includegraphics[width=.45\linewidth]{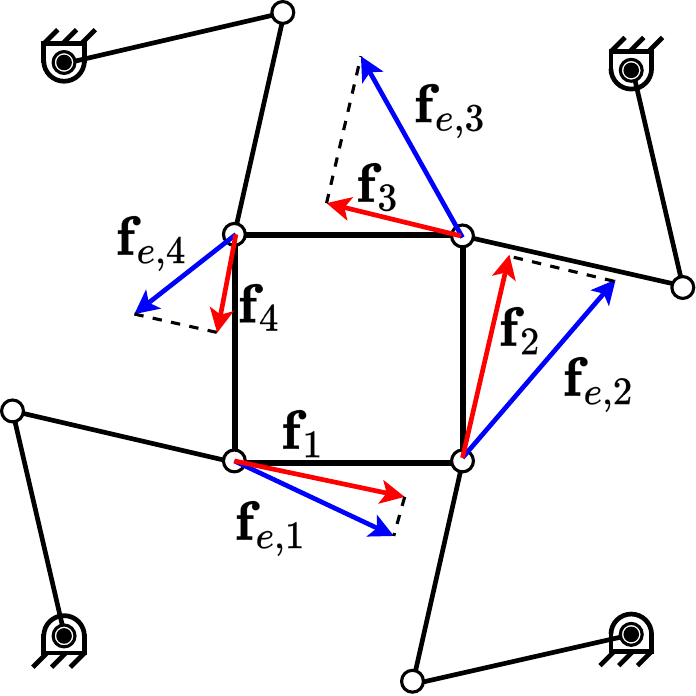}%
        \label{subfig:Planar4RRR-LeastSquares}%
    } \quad
    \subfloat[4-\underline{RR}R manipulator with statically determined legs. Eq. (\ref{eq:jointTorque2AppliedWrench}) has a unique solution and the legs are able to generate the equilibrating force distribution $\mb{f}_e$.]{%
        \includegraphics[width=.45\linewidth]{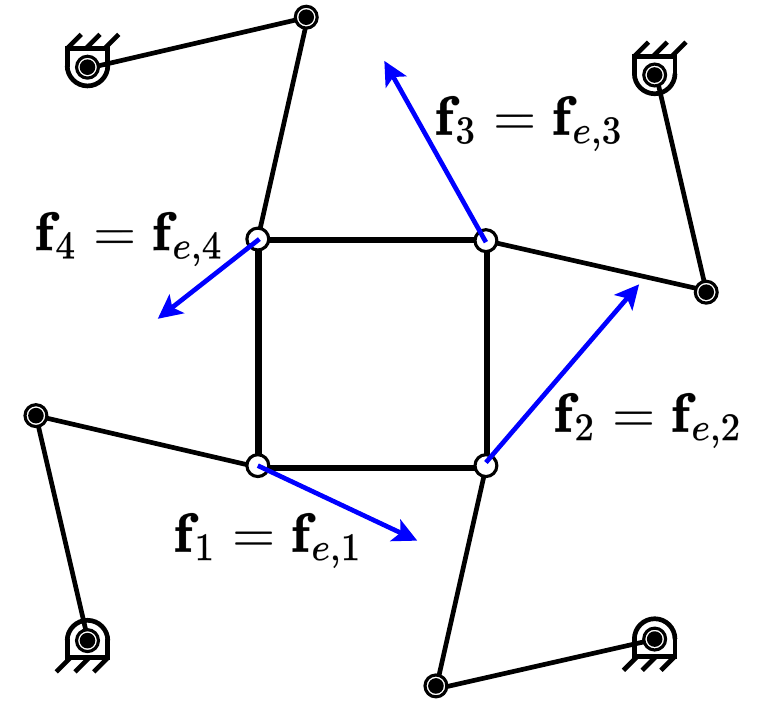}%
        \label{subfig:Planar4RRR-Exact}%
    }
    \caption{Effect of the static determinacy of the legs of the manipulator on the solution returned by inverting (\ref{eq:jointTorque2AppliedWrench}) for a desired resultant wrench $\mb{h}_o = \begin{bmatrix}
        f_x & f_y & m_z
    \end{bmatrix}^T = \begin{bmatrix}
        0.091 & 0.182 & 0.091
    \end{bmatrix}^T$. The static determinacy of the legs determines whether the legs can produce the correct equilibrating force distribution $\mb{f}_e$.}
    \label{fig:Planar4RRR}
\end{figure}

Figure \ref{fig:Planar4RRR} illustrates the physical significance of inverting (\ref{eq:jointTorque2AppliedWrench}) when the legs can generate the required wrenches versus when they cannot. Both manipulators have actuation redundancy, but in Fig \ref{subfig:Planar4RRR-LeastSquares}, the 4-\underline{R}RR manipulator has one actuator per leg and each leg produces a force with two components. This is an example of Case 2 where the legs are statically underdetermined. If we invert (\ref{eq:jointTorque2AppliedWrench}), the joint torque vector returned will be that which minimizes the error in the linear system, but will not be an exact solution. The physical significance of this result is that these joint torques will generate the forces that are the projections of the \emph{desired} solution onto the directions in which the legs \emph{can} produce forces. For this case, it means projecting every vector $\mb{f}_{e,i}$ onto the line passing through the distal link, since this is the direction of the forces produced by the legs. 

Conversely, in Fig. \ref{subfig:Planar4RRR-Exact}, we have added an additional actuator to each leg to create a 4-\underline{RR}R manipulator. Each leg now has two actuator and produces a force with two components, making them statically determined. The legs are therefore able to produce forces in the required directions, and a valid solution can be found for the equilibrating joint torque vector. 

\begin{figure}
    \centering    \includegraphics[width=0.4\linewidth]{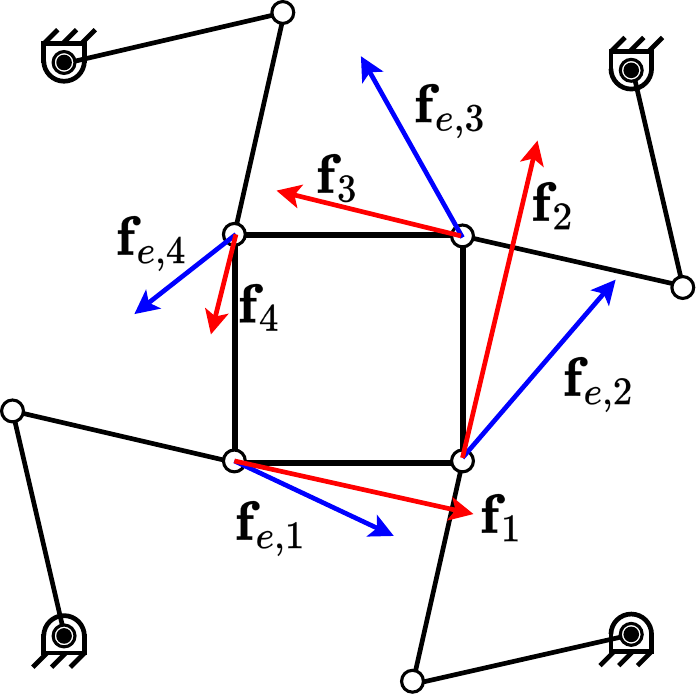}
    \caption{Solution returned by the minimum-norm solution for $\boldsymbol{\tau}$ for the resultant wrench $\mb{h}_o = \begin{bmatrix}
        f_x & f_y & m_z
    \end{bmatrix}^T = \begin{bmatrix}
        0.091 & 0.182 & 0.091
    \end{bmatrix}^T$ for the planar 4-\underline{R}RR manipulator. The correct equilibrating force distribution $\mb{f}_e$ is shown in blue and the actual forces $\mb{f}$ generated by the legs of the manipulator are shown in red. }
    \label{fig:Planar4RRR-MNS}
\end{figure}

Note that it is still possible to synthesize joint torque vectors that generate the required resultant wrench even when the legs are statically underdetermined. Fig. \ref{fig:Planar4RRR-MNS} shows the forces produced by the minimum-norm solution for the joint torques that corresponds to the same resultant wrench used in Fig. \ref{fig:Planar4RRR}. Evidently, this is neither a valid equilibrating joint torque vector nor a valid manipulating joint torque vector, but taking the sum of the forces does produce the correct resultant wrench. 

\subsection{Obtaining the Mapping Matrices}
\begin{figure}
    \centering
    \includegraphics[width=0.3\linewidth]{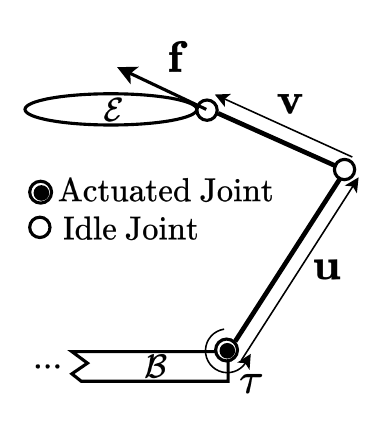}
    \caption{Force transmitted to the end-effector by a planar \underline{R}RR leg.}
    \label{fig:SingleLegExample}
\end{figure}

To better illustrate the physical significance of this mapping, let us determine the relevant quantities for one actuator in a simplified example. Consider a planar \underline{R}RR leg with one actuator fixed to the base $\mathcal{B}$ of a parallel manipulator as shown in Fig. \ref{fig:SingleLegExample}. Vectors $\mb{u}$ and $\mb{v}$ run along the proximal and distal links of the leg, respectively, and the effect of joint torque $\tau$ is felt at the end-effector $\mathcal{E}$ as a pure force $\mb{f}$ which is collinear with the distal link (i.e., in the direction of $\mb{v}$).

\begin{figure}
\centering
    \subfloat[Proximal link.]{%
        \includegraphics[width=.4\linewidth]{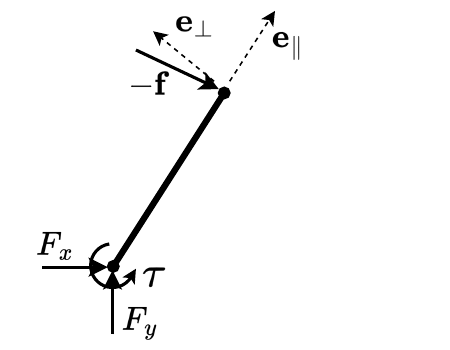}%
        \label{subfig:FreeBodyDiagram-Proximal}%
    } \quad
    \subfloat[Distal link.]{%
        \includegraphics[width=.4\linewidth]{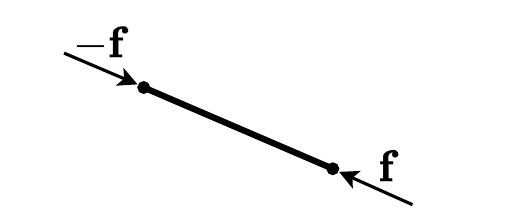}%
        \label{subfig:FreeBodyDiagram-Distal}%
    }
    \caption{Free-body diagrams of the links of the \underline{R}RR parallel robot leg.}
    \label{fig:FreeBodyDiagrams}
\end{figure}

The free-body diagrams of the proximal and distal links under these loading conditions are shown in Fig. \ref{fig:FreeBodyDiagrams}. Vectors $\mb{e}_{\parallel}$ and $\mb{e}_{\perp}$ are unit vectors that are parallel and perpendicular to the proximal link, respectively. If the leg applies a force $\mb{f}$ to the end-effector, then the end-effector applies an equal and opposite force $-\mb{f}$ to the leg. However, the joint torque $\tau$ need only resist force components applied at the end of the proximal link which are perpendicular to $\mb{u}$ (i.e., in the direction of $\mb{e}_{\perp}$). Any component parallel to $\mb{u}$ is compensated by a reaction force at the base and requires no effort from $\tau$ to counteract.  

Force $\mb{f}$ can be expressed as 

\begin{equation}
    \mb{f} = w \mb{v}
    \label{eq:forceFromTransmissionWeight}
\end{equation}
where $w$ is the transmission weight which modulates the intensity of the force along vector $\mb{v}$ in accordance with the magnitude of the joint torque $\tau$. Hence, by expressing the transmission weight $w$ as a function of $\tau$, we can express the applied force $\mb{f}$ as a function of $\tau$ as well.

The magnitude of force $-\mb{f}$ in the direction of $\mb{e}_{\perp}$ is obtained as

\begin{equation}
    ||\mb{f}_{\perp}||_2 = \left| -\mb{f}^T \mb{e}_{\perp} \right| =  \left|-\mb{f}^T \mb{E} \frac{\mb{u}}{|| \mb{u} ||}\right|
    \label{eq:perpForceComponent}
\end{equation}
with

\begin{equation}
    \mb{E} = \begin{bmatrix}
        0 & -1 \\ 1 & 0
    \end{bmatrix}.
    \label{eq:matrixE}
\end{equation}
To counteract the applied force, $\tau$ must generate an equal and opposite force along $\mb{e}_{\perp}$. A torque $\tau$ generates a force of magnitude $F$ at the end of a lever of length $l$ given by 

\begin{equation}
    F = \frac{\tau}{l}.
    \label{eq:forceFromTorque}
\end{equation}
Hence, for the forces to cancel, we must have

\begin{equation}
    \frac{\tau}{|| \mb{u} ||_2} = - \left( -\mb{f}^T\mb{E} \frac{\mb{u}}{|| \mb{u} ||_2} \right)
    \label{eq:forceComponentEquivalence}
\end{equation}
since the length of the lever for $\tau$ in this case is $||\mb{u}||_2$. Finally, we substitute (\ref{eq:forceFromTransmissionWeight}) into (\ref{eq:forceComponentEquivalence}) to obtain

\begin{equation}
    \tau = w \mb{v}^T \mb{E} \mb{u},
\end{equation}
which leads to our desired result

\begin{equation}
    w = \frac{\tau}{\mb{v}^T \mb{E} \mb{u}}.
    \label{eq:transmissionWeight}
\end{equation}

Eq. (\ref{eq:transmissionWeight}) can be used to convert $\tau$ to the magnitude of the force that it generates. The force generated by a given joint torque $\tau$ is therefore written as

\begin{equation}
    \mb{f} = \frac{\tau}{\mb{v}^T \mb{E} \mb{u}} \mb{v}
    \label{eq:forceFromTorqueExample}
\end{equation}

If we had a manipulator with multiple legs like the one in this example and we sought to define the mapping (\ref{eq:jointTorque2AppliedWrench}), the element in $\mb{K}^{-T}$ which corresponds to the $i$-th actuated joint would be given by $(\mb{v}_i^T \mb{E} \mb{u}_i)^{-1}$ and the force direction to be used in basis matrix $\mb{B}$ would be $\mb{v}_i$.

\section{The Problem with Existing Methods: Algebraic vs. Physical Norms}
In the literature, some authors have claimed that the minimum-norm solution for $\boldsymbol{\tau}$ for a given resultant wrench $\mb{h}_o$ does not generate internal loads \cite{Nokleby2005, Zibil2007, Garg2009}. The reasons why this is an erroneous claim were given in Section \ref{sec:RelatedWork}, and this is further supported by considering the mapping of the joint efforts to applied wrenches defined in (\ref{eq:jointTorque2AppliedWrench}).

Crucially, the method proposed in \cite{Nokleby2005} for eliminating interaction forces overlooks the fact that the Euclidean norm of a vector of parameters which define distances along basis vectors is not the same as the Euclidean norm of the actual spatial vector which they define; i.e., the Euclidean norm of the vector $\boldsymbol{\tau}$ which defines the applied wrenches differs from the Euclidean norm of the applied wrench vector itself. 

Consider the following, the Euclidean norm of an arbitrary $n$-dimensional vector $\mathbf{x} = \begin{bmatrix}
    x_1 & \ldots & x_n
\end{bmatrix}^T$ is defined as 

\begin{equation}
    ||\mathbf{x}||_2 = \sqrt{x_1^2 + x_2^2 + \cdots + x_n^2}.
    \label{eq:2norm}
\end{equation}
The unweighted Moore-Penrose pseudo-inverse returns the unique solution to an underdetermined linear system of equations with the smallest Euclidean norm, which is why (\ref{eq:KWMinNormSolution}) returns the minimum-norm solution for $\mathbf{f}$ for a given wrench vector $\mathbf{h}_o$ (i.e., $\text{min} (\mathbf{f}^T \mathbf{f})$). However, (\ref{eq:2norm}) only holds when the components $x_i$ define distances along orthonormal axes. Much like how a 3-dimensional vector in Cartesian space is defined by its $x$, $y$, and $z$ components along the $x$-, $y$-, and $z$-axes, which are mutually orthogonal.

\begin{figure}
    \centering
    \includegraphics[width=0.6\linewidth]{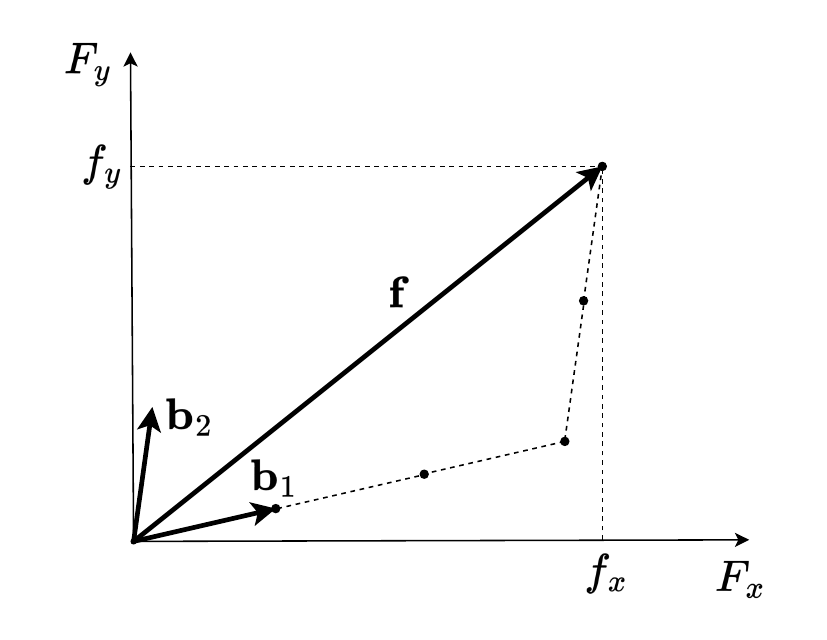}
    \caption{The linear combination of the basis vectors $\mathbf{b}_1$ and $\mathbf{b}_2$ defines the resultant force $\mathbf{f}$.}
    \label{fig:ForceComponents}
\end{figure}

To illustrate the effect of non-orthogonal basis vectors on the resulting vector's norm, consider the following example. Let a general force vector $\mathbf{f} = \begin{bmatrix}
    f_x & f_y
\end{bmatrix}^T$ be constructed from two non-orthogonal basis vectors $\mathbf{b}_1$ and $\mathbf{b}_2$ as shown in Fig. \ref{fig:ForceComponents}. Vector $\mathbf{f}$ may be written as

\begin{equation}
    \mathbf{f} = q_1 \mathbf{b}_1 + q_2 \mathbf{b}_2 = \begin{bmatrix}
        \mathbf{b}_1 & \mathbf{b}_2
    \end{bmatrix} \begin{bmatrix}
        q_1 \\ q_2
    \end{bmatrix} = \mathbf{B} \mathbf{q}.
\end{equation}

\noindent
If we choose arbitrarily $\mathbf{q} = \begin{bmatrix}
    3 & 2
\end{bmatrix}^T$, we obtain the vector $\mathbf{f}$ shown in Fig. \ref{fig:ForceComponents}. 

It is then evident that the Euclidean norm of $\mathbf{f}$ is not obtained from the sum of the squares of the components of $\mathbf{q}$ as shown in (\ref{eq:2norm}). Rather, it is obtained with

\begin{equation}
    ||\mathbf{f}||_2^2 = f_x^2 + f_y^2 = q_1^2 + q_2^2 + 2 q_1 q_2 \mathbf{b}_1 \cdot \mathbf{b}_2.
\end{equation}
Hence, when finding the Euclidean norm of a vector determined by parameters which define distances along basis vectors, it is imperative to consider the relative angles between the basis vectors (i.e., the term $2 q_1 q_2 \mathbf{b}_1 \cdot \mathbf{b}_2$). This issue can be resolved by weighting the norm of the parameter vector using a metric tensor.

\section{The Generalized Method: A Metric Tensor Approach}
\label{sec:GeneralizedMethod}
We may use (\ref{eq:jointTorque2AppliedWrench}) to derive a generalized method which accounts for the indirect transmission of joint torques to the end-effector. By substituting (\ref{eq:jointTorque2AppliedWrench}) into (\ref{eq:InteractionForceOptProb}), we can write the optimization problem which defines the equilibrating force distribution as a function of $\boldsymbol{\tau}$. We obtain

\begin{equation}
    \begin{aligned}
        \min& \qquad Z = \frac{1}{2} \mb{f}^T \mb{f} = \frac{1}{2} \boldsymbol{\tau}^T (\mb{K}^{-1} \mb{B}^T \mb{B} \mb{K}^{-T}) \boldsymbol{\tau}\\ 
        \text{subject to}& \qquad \mb{h}_o = \mb{G} \mb{f} = \mb{G} \mb{B} \mb{K}^{-T} \boldsymbol{\tau} = \mb{J}^T \mb{K}^{-T} \boldsymbol{\tau}.
        \label{eq:ModifiedInteractionForceOptProb}
    \end{aligned}
\end{equation}
Hence, we have transformed an unweighted quadratic minimization problem into a weighted one. The weighting matrix $\mb{W}_e = \mb{K}^{-1} \mb{B}^T \mb{B} \mb{K}^{-T}$ is the Riemannian Metric Tensor which accounts for the non-orthogonality of the wrench directions of the actuators. 

An explicit solution for $\boldsymbol{\tau}$ which balances a desired wrench without generating interaction forces can be easily derived using the Lagrange multiplier formulation. We write the Lagrangian as 

\begin{equation}
    \mathcal{L}(\boldsymbol{\tau}, \boldsymbol{\lambda}) = \frac{1}{2} \boldsymbol{\tau}^T (\mb{K}^{-1} \mb{B}^T \mb{B} \mb{K}^{-T}) \boldsymbol{\tau} + (\mb{h}_o - \mb{J}^T \mb{K}^{-T} \boldsymbol{\tau})^T \boldsymbol{\lambda}.
\end{equation}
where $\boldsymbol{\lambda}$ is the vector of Lagrange multipliers. Taking the partial derivatives, we obtain 

\begin{align}
    \frac{\delta \mathcal{L}}{\delta \boldsymbol{\tau}} &= (\mb{K}^{-1} \mb{B}^T \mb{B} \mb{K}^{-T}) \boldsymbol{\tau} - (\mb{J}^T \mb{K}^{-T})^T \boldsymbol{\lambda} = \mb{0}, \label{eq:LagrangianPD1}\\
    \frac{\delta \mathcal{L}}{\delta \boldsymbol{\lambda}} &= \mb{h}_o - \mb{J}^T \mb{K}^{-T} \boldsymbol{\tau} = \mb{0} \label{eq:LagrangianPD2}
\end{align}
From (\ref{eq:LagrangianPD1}), we can write

\begin{equation}
    \boldsymbol{\tau} = (\mb{K}^{-1} \mb{B}^T \mb{B} \mb{K}^{-T})^{-1} (\mb{J}^T \mb{K}^{-T})^T \boldsymbol{\lambda}.
    \label{eq:tauFromLagrangeMultiplier}
\end{equation}
This can be substituted into (\ref{eq:LagrangianPD2}) to solve for $\boldsymbol{\lambda}$ as 

\begin{equation}
    \boldsymbol{\lambda} = \left[ \mb{J}^T \mb{K}^{-T} (\mb{K}^{-1} \mb{B}^T \mb{B} \mb{K}^{-T})^{-1} (\mb{J}^T \mb{K}^{-T})^T \right]^{-1} \mb{h}_o.
    \label{eq:LagrangeMultiplier}
\end{equation}
Finally, we substitute (\ref{eq:LagrangeMultiplier}) into (\ref{eq:tauFromLagrangeMultiplier}) to obtain the solution 

\begin{equation}
    \boldsymbol{\tau}_e = (\mb{J}^T \mb{K}^{-T})^{\dagger}_{\mb{W}_e} \mb{h}_o,
    \label{eq:EquilibratingJointTorques}
\end{equation}
where $\boldsymbol{\tau}_e$ is the \emph{equilibrating joint torque vector} and 

\begin{equation}
    (\mb{J}^T \mb{K}^{-T})^{\dagger}_{\mb{W}_e} = \mb{W}_e^{\dagger} (\mb{J}^T \mb{K}^{-T})^T \left[ \mb{J}^T \mb{K}^{-T} \mb{W}_e^{\dagger} (\mb{J}^T \mb{K}^{-T})^T \right]^{-1}
\end{equation}
is the weighted Moore-Penrose pseudo-inverse of $\mb{J}^T \mb{K}^{-T}$ with weighting matrix $\mb{W}_e = \mb{K}^{-1} \mb{B}^T \mb{B} \mb{K}^{-T}$. The forces generated by $\boldsymbol{\tau}_e$ generate the desired resultant wrench without generating interaction forces. 

Since the optimization problem (\ref{eq:InternalLoadOptProb}) which characterizes the manipulating wrench distribution is analogous to that of the equilibrating force distribution, we can obtain a joint torque vector which does not generate internal loads in much the same way. If we write (\ref{eq:InternalLoadOptProb}) as a function of $\boldsymbol{\tau}$, we obtain

\begin{equation}
    \begin{aligned}
        \min& \qquad Z =  \mb{h}^T \mb{M}^{-1} \mb{h} = \boldsymbol{\tau}^T (\mb{K}^{-1} \mb{B}^T \mb{M}^{-1} \mb{B} \mb{K}^{-T}) \boldsymbol{\tau}\\ 
        \text{subject to}& \qquad \mb{h}_o = \mb{G} \mb{h} = \mb{G} \mb{B} \mb{K}^{-T} \boldsymbol{\tau} = \mb{J}^T \mb{K}^{-T} \boldsymbol{\tau}.
        \label{eq:ModifiedInternalLoadOptProb}
    \end{aligned}
\end{equation}
Then, using the Lagrangian, we obtain the solution 

\begin{equation}
    \boldsymbol{\tau}_m = (\mb{J}^T \mb{K}^{-T})^+_{\mb{W}_m} \mb{h}_o
    \label{eq:ManipulatingJointTorques}
\end{equation}
where $\boldsymbol{\tau}_m$ is the \emph{manipulating joint torque vector} and 

\begin{equation}
    (\mb{J}^T \mb{K}^{-T})^+_{\mb{W}_m} = \mb{W}_m^{\dagger} (\mb{J}^T \mb{K}^{-T})^T \left[ \mb{J}^T \mb{K}^{-T} \mb{W}_m^{\dagger} (\mb{J}^T \mb{K}^{-T})^T \right]^{-1}
\end{equation}
is the weighted parametrized Moore-Penrose pseudo-inverse of $\mb{J}^T \mb{K}^{-T}$ with weighting matrix $\mb{W}_m = \mb{K}^{-1} \mb{B}^T  \mb{M}^{-1}\mb{B} \mb{K}^{-T}$. The forces generated by $\boldsymbol{\tau}_m$ generate the desired resultant wrench without generating internal loads.

Note that we are taking the pseudo-inverse of the weighting matrices $\mb{W}_e$ and $\mb{W}_m$ to account for scenarios when a manipulator has redundantly-actuated legs and matrix $\mb{B}$ becomes a wide matrix. For a solution to exist for the equilibrating and manipulating joint torque vectors, the following conditions must be met:

\begin{enumerate}
    \item Jacobian matrix $\mb{K}$ must have full rank (i.e., manipulator is not in a type I singularity).
    \item Jacobian matrix $\mb{J}$ must have full rank (i.e., manipulator is not in a type II singularity).
    \item Basis matrix $\mb{B}$ must have full row rank.
    \item Mass matrix $\mb{M}$ must be positive definite.
\end{enumerate}

\section{Case Studies}
This section presents two case studies in which we apply our novel methods to the analysis of parallel manipulators with actuation redundancy. The first case study serves to highlight some critical oversights in past work and provide corrections, while the second case study serves to show the generality of our methods by analyzing a spatial manipulator where the legs can apply both forces and torques to the end-effector. 

\subsection{Case Study 1: 3-\underline{RR}R Planar Parallel Manipulator}
Let us now use the generalized method presented in the previous section to demonstrate how methods such as the one proposed by Kumar and Waldron in \cite{Kumar1988} have been erroneously applied to parallel manipulators in the literature. 

In this section, we give an example where Kumar and Waldron's method is used to analyze a 3-\underline{RR}R planar parallel manipulator with actuation redundancy. We show that the method used by the authors and our generalized method return different results, and that only our generalized method returns force sets free of interaction forces.

\subsubsection{Manipulator Architecture}
\begin{figure}
    \centering
    \includegraphics[width=0.7\linewidth]{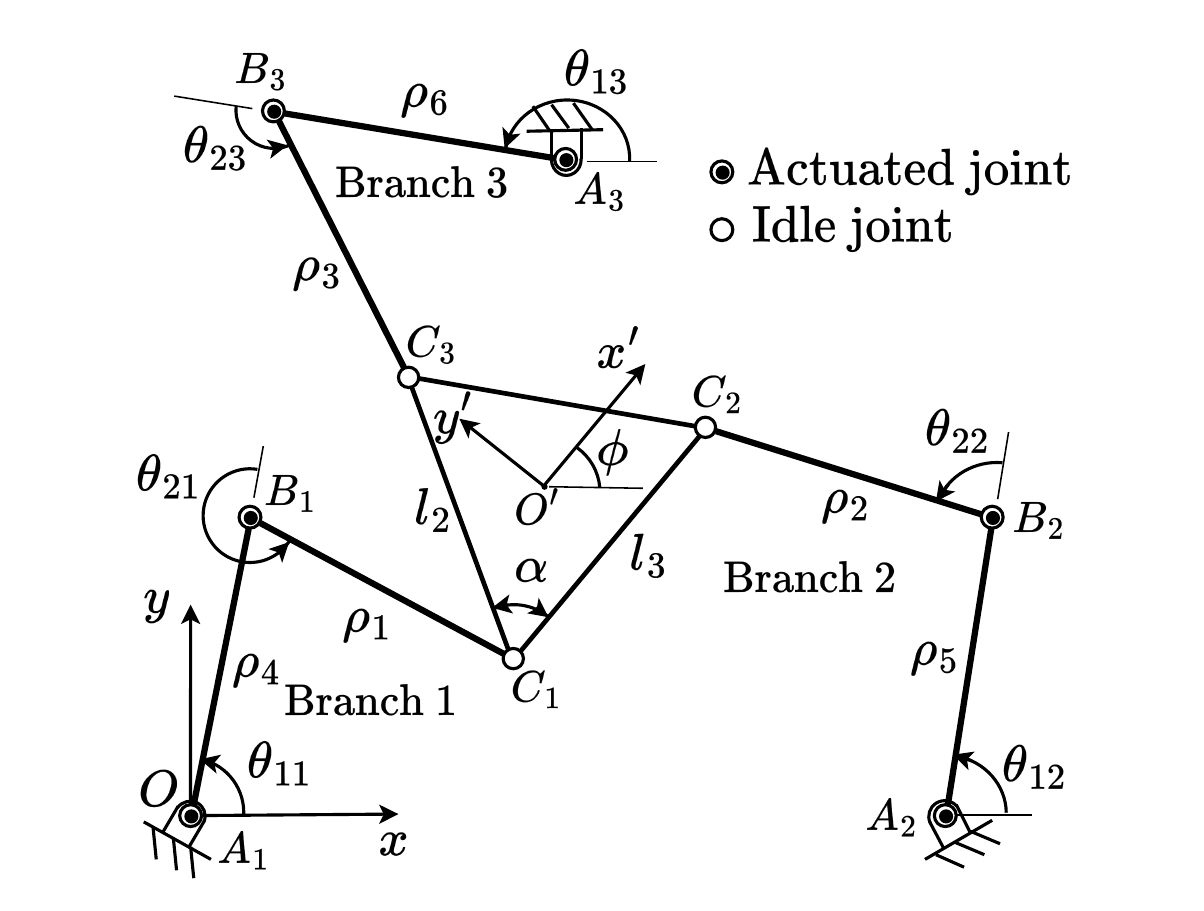}
    \caption{3-\underline{RR}R planar parallel manipulator with actuation redundancy. Adapted from \cite{Nokleby2005}.}
    \label{fig:3RRR}
\end{figure}

The chosen example is taken from \cite{Nokleby2005}. Fig. \ref{fig:3RRR} shows a schematic of the manipulator architecture. The geometry of the manipulator is defined by the link lengths $\rho_1, ..., \rho_6$ and the end-effector geometry parameters $l_2$, $l_3$, and $\alpha$. The pose of the end-effector is defined by $\mb{x} = \begin{bmatrix}
    x & y & \phi
\end{bmatrix}^T$ where $x$ and $y$ define the coordinates of the origin $O'$ of the mobile frame in the base frame and angle $\phi$ defines the angle of the end-effector with respect to the $x$-axis of the base frame. Three legs, also called branches, with revolute actuators are mounted in parallel to the end-effector, which is assumed to be a rigid body. For this architecture, each leg has 2 actuators and can generate a force vector in any direction in the plane. It is therefore possible to find equilibrating and manipulating joint torque vectors.

Each branch transmits a pure force to the end-effector. To remain consistent with the notation in \cite{Nokleby2005}, the subscript $ji$ designates the $j$-th actuated joint of the $i$-th branch. Let vectors $\mb{a}_i$, $\mb{b}_i$, and $\mb{c}_i$ define the locations of points $A_i$, $B_i$ and $C_i$ with respect to the origin of the base frame, respectively. We then let vector $\mb{r}_i$ define the location of point $C_i$ with respect to $O'$, expressed in the base frame, and we define vectors $\mb{u}_i$ and $\mb{v}_i$ as 

\begin{align}
    \mb{u}_i = \mb{b}_i - \mb{a}_i, \\
    \mb{v}_i = \mb{c}_i - \mb{b}_i,
\end{align}
to represent the proximal and distal links of each leg, respectively.

In \cite{Nokleby2005}, the authors analyze both a non-redundant and redundantly-actuated version of the manipulator. Here, we only analyze the redundantly-actuated architecture where the first two revolute joints of each branch are actuated. Note that our notation differs from the one used in \cite{Nokleby2005}.
The manipulator has six actuators which constrain three wrench components. The wrench equilibrium equation is written as

\begin{equation}
    \mb{h}_o = \mb{J}^T \mb{K}^{-T} \boldsymbol{\tau}
\end{equation}

\noindent
where $\mathbf{h}_o = \begin{bmatrix}
    f_x & f_y & m_z
\end{bmatrix}^T$ is the resultant wrench at the end-effector, $\mb{J}^T$ is a $3 \times 6$ matrix whose $j$-th column is the wrench direction (able to be defined by only 3 components since it lies in the plane) of the $j$-th actuator, and matrix $\mb{K}^{-T}$ is a $6 \times 6$ diagonal matrix of transmission weights. $\mb{J}$ and $\mb{K}$ are written as 

\begin{align}
    \mb{J} &= \begin{bmatrix}
        \mb{v}_{1}^T & \mb{v}_1^T \mb{E} \mb{r}_1 \\
        (\mb{u}_1 + \mb{v}_1)^T & (\mb{u}_1 + \mb{v}_1)^T \mb{E} \mb{r}_1\\
        \mb{v}_{2}^T & \mb{v}_2^T \mb{E} \mb{r}_2 \\ 
        (\mb{u}_2 + \mb{v}_2)^T & (\mb{u}_2 + \mb{v}_2)^T \mb{E} \mb{r}_2\\
        \mb{v}_{3}^T & \mb{v}_3^T \mb{E} \mb{r}_3\\
        (\mb{u}_3 + \mb{v}_3)^T & (\mb{u}_3 + \mb{v}_3)^T \mb{E} \mb{r}_3
    \end{bmatrix}, \label{eq:3RRR-J}\\
    \mb{K} &= \begin{bmatrix}
        \mb{K}_1 & \mb{0}_2 & \mb{0}_2 \\
        \mb{0}_2 & \mb{K}_2 & \mb{0}_2 \\
        \mb{0}_2 & \mb{0}_2 & \mb{K}_3
    \end{bmatrix}
    \label{eq:3RRR-K}
\end{align}
where 

\begin{equation}
    \mb{K}_i = \begin{bmatrix}
        \mb{v}_i^T \mb{E} \mb{u}_i & 0 \\ 
        0 & ||\mb{u}_i|| ||\mb{v}_i|| \sin(\theta'_{2i})
    \end{bmatrix}
\end{equation}
with $\theta'_{2i}$ being the inside angle between $\mb{u}_i$ and $\mb{v}_i$ and $\mb{E}$ is defined as in (\ref{eq:matrixE}).

\begin{figure}
    \centering
    \includegraphics[width=0.5\linewidth]{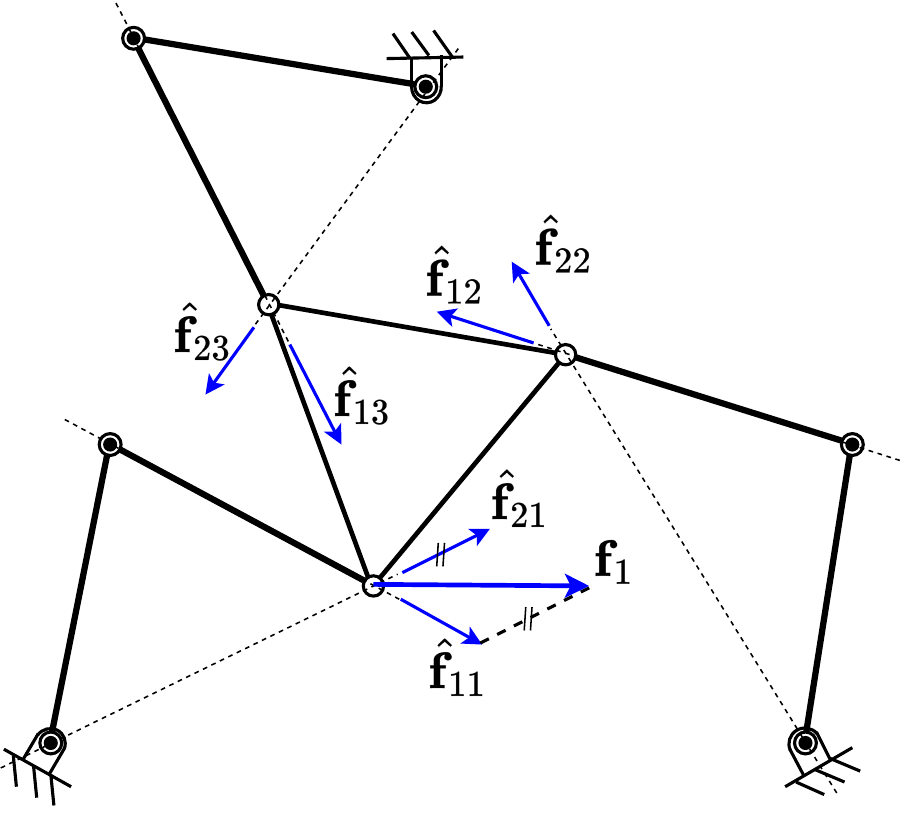}
    \caption{Screw axes of the forces applied to end-effector of the 3-\underline{RR}R planar parallel manipulator by the actuated joints (forces not to scale).}
    \label{fig:3RRR-Screws}
\end{figure}

Since the form of matrices $\mb{J}$ and $\mb{K}$ that we provide differ from those given in \cite{Nokleby2005}, we have included their derivation in Appendix A (both forms are equivalent). The forces generated by each actuator are shown at their respective application points in Fig. \ref{fig:3RRR-Screws} where $\hat{\mb{f}}_{ji}$ is the force produced by a unit torque applied by the $j$-th actuator on the $i$-th leg. 

The resultant force applied to the end-effector by each leg is a linear combination of the force directions of the joints on that leg with their corresponding transmission weights as coefficients. The force $\mathbf{f}_{i}$ applied by leg $i$ is obtained with

\begin{equation}
    \mb{f}_i = \frac{\tau_{1i}}{k_{i,11}} \hat{\mb{f}}_{1i} + \frac{\tau_{2i}}{k_{i,22}} \hat{\mb{f}}_{2i} \quad i = 1, 2, 3
\end{equation}
where $\tau_{1i}$ and $\tau_{2i}$ are the torques applied by actuated joints 1 and 2 on leg $i$ and $k_{i,ab}$ is element $ab$ of matrix $\mb{K}_i$. An example of how the forces produced by the actuators on a given leg sum to produce the resultant force applied to the end-effector by that leg is also shown for leg 1 in Fig. \ref{fig:3RRR-Screws}.

\subsubsection{Methodology}
The authors of \cite{Nokleby2005} present two feasible force polygons with a prescribed moment of $m_z = 0$ for the redundantly-actuated manipulator. The first polygon is obtained using the \emph{scaling factor method} which they propose in the same paper. The method first finds the minimum-norm joint torque vector required to generate a unit wrench in a given direction with

\begin{equation}
    \hat{\boldsymbol{\tau}} = (\mb{J}^T \mb{K}^{-T})^{\dagger} \hat{\mb{h}}
    \label{eq:minNormJointTorqueUnitWrench}
\end{equation}

\noindent
where $\hat{\mb{h}}$ is the unit resultant wrench and $(\mb{J}^T \mb{K}^{-T})^{\dagger}$ is the unweighted Moore-Penrose pseudo-inverse of $\mb{J}^T \mb{K}^{-T}$ which is defined as 

\begin{equation}
    (\mb{J}^T \mb{K}^{-T})^{\dagger} = (\mb{J}^T \mb{K}^{-T})^{T} \left[ (\mb{J}^T \mb{K}^{-T}) (\mb{J}^T \mb{K}^{-T})^{T} \right]^{-1}.
\end{equation}

\noindent
They then determine the scaling factor $S$, which is the maximum factor by which all joint torques can be scaled while remaining within their allowable range. The maximum wrench that can be applied in the direction of $\hat{\mb{h}}$ is then 

\begin{equation}
    \mathbf{h}_o = S \mb{J}^T \mb{K}^{-T} \hat{\boldsymbol{\tau}}
\end{equation}

\noindent
The polygon is constructed by repeating this process for every direction in the plane.

The second polygon represents the true maximum feasible force set that can be achieved by the manipulator with all actuators being exploited to their maximum potential. In \cite{Nokleby2005}, the authors use an optimization-based method to determine this polygon. Here, we take an alternate approach of first defining the feasible set of joint torques in the joint space and then applying the linear transformation defined by $\mb{J}^T \mb{K}^{-T}$ to map this feasible set to the task space. The feasible set for $\boldsymbol{\tau}$ is defined by

\begin{equation}
    \mathcal{T} = \{ \boldsymbol{\tau} \in \mathbb{R}^6 \mid ||\boldsymbol{\tau}||_{\infty} \leq \tau_{max} \}
\end{equation}

\noindent
where $||\cdot||_{\infty}$ represents the $\infty$-norm of the argument $(\cdot)$ and $\tau_{max}$ is the maximum allowable torque value. The polyhedron of feasible wrenches is then easily obtained by mapping this feasible set to the task space. The set of feasible wrenches $\mathcal{W}$ is therefore given by

\begin{equation}
    \mathcal{W} = \{ \mathbf{h}_o \in \mathbb{R}^3 \mid \mathbf{h}_o = \mb{J}^T \mb{K}^{-T} \boldsymbol{\tau}, \ \boldsymbol{\tau} \in \mathcal{T} \}.
\end{equation}

\noindent
The result is a polyhedron (more specifically a zonohedron, since $\mathcal{W}$ is centrally symmetric) in the task space. The polygon of feasible forces for a prescribed moment of $m_z = 0$ is obtained from the intersection of $\mathcal{W}$ and the plane defined by $m_z = 0$. 

Let us now determine the feasible force polygon for a prescribed moment of $m_z = 0$ using the generalized method proposed in Section \ref{sec:GeneralizedMethod}. To do so, we must determine matrix $\mb{B}$ to be used in (\ref{eq:jointTorque2AppliedWrench}). For this manipulator, it is written as

\begin{equation}
    \mathbf{B} = \begin{bmatrix}
        \mathbf{v}_{1} & \mb{u}_1 + \mb{v}_1 & \mathbf{0}_{2 \times 1} & \mathbf{0}_{2 \times 1} & \mathbf{0}_{2 \times 1} & \mathbf{0}_{2 \times 1} \\
        \mathbf{0}_{2 \times 1} & \mathbf{0}_{2 \times 1} & \mathbf{v}_{2} & \mb{u}_2 + \mb{v}_2 & \mathbf{0}_{2 \times 1} & \mathbf{0}_{2 \times 1} \\
        \mathbf{0}_{2 \times 1} & \mathbf{0}_{2 \times 1} & \mathbf{0}_{2 \times 1} & \mathbf{0}_{2 \times 1} & \mathbf{v}_{3} & \mb{u}_3 + \mb{v}_3
    \end{bmatrix}.
\end{equation}
This matrix can then be used to calculate the weighted Moore-Penrose pseudo-inverse of $\mb{J}^T \mb{K}^{-T}$ and apply the scaling factor method as described in \cite{Nokleby2005} to find the new feasible force polygon. 

Note that, since the end-effector is an equilateral triangle, the virtual mass distribution which satisfies constraints (\ref{eq:virtualMassSum})-(\ref{eq:inertiaProportionality}) is $m^*_1 = m^*_2 = m^*_3$. In addition, the legs can only apply pure forces, so we also have $\mb{I}^*_1 = \mb{I}^*_2 = \mb{I}^*_3 = \mb{0}_3$. The inertia matrix $\mb{M}$ of the system is therefore proportional to the identity matrix, and the equilibrating force distribution and the manipulating wrench distribution (which in this case can only include pure forces) are identical \cite{Flight2026}.

\subsubsection{Results}
In \cite{Nokleby2005}, the link lengths and the end-effector edge lengths are all equal to 0.200 m (i.e. $\rho_1 = ... = \rho_6 = l_1 = l_2 = 0.200$ m), and $\alpha = 60$\degree. The bases of the branches are 0.500 m apart. The pose is set to $\mb{x} = \begin{bmatrix}
    0.250 & 0.144 & 0
\end{bmatrix}^T$, the joint positions at this pose are $ \{ \theta_{11}, \theta_{21}, \theta_{12}, \theta_{22}, \theta_{13}, \theta_{23} \} = \{94.34, -128.68, 214.34, -128.68, 334.34, -128.68 \}^{\degree}$ and $\tau_{max}$ is chosen to be 4.2 Nm.

\begin{figure}
    \centering
    \begin{overpic}[width=0.5\linewidth,tics=10]{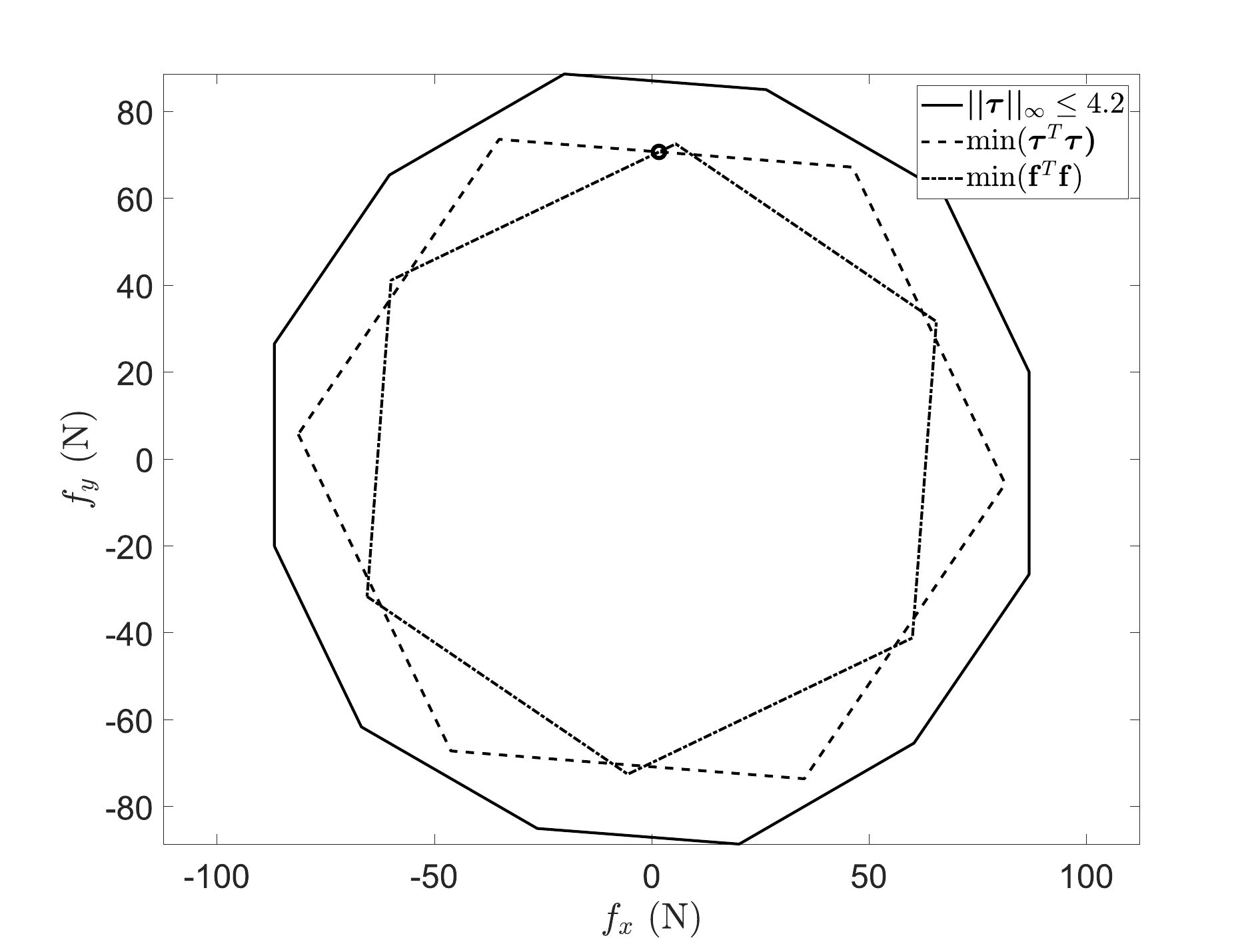}
        \put(4, 75){\rotatebox{90}{%
                \colorbox{white}{$f_y$ (N)}%
            }} 
        \put(110,1){\colorbox{white}{$f_x$ (N)}}
        \put(147, 144){%
            \fcolorbox{black}{white}{%
                \renewcommand{\arraystretch}{1}%
                \begin{tabular}{@{\hspace{-0.2em}}c @{\hspace{0.3em}}l@{\hspace{-0.2em}}}
                    
                    \rule[0.6ex]{2em}{1.5pt} & $\|\boldsymbol{\tau}\|_{\infty} \leq 4.2$ \\
                    
                    \rule[0.6ex]{0.5em}{1.5pt}\hspace{0.25em}%
                    \rule[0.6ex]{0.5em}{1.5pt}\hspace{0.25em}%
                    \rule[0.6ex]{0.5em}{1.5pt} & $\min(\boldsymbol{\tau}^T \boldsymbol{\tau})$ \\
                    
                    \rule[0.6ex]{0.7em}{1.5pt}\hspace{0.2em}%
                    \rule[0.6ex]{0.2em}{1.5pt}\hspace{0.2em}%
                    \rule[0.6ex]{0.7em}{1.5pt}\hspace{0.2em} & $\min(\mathbf{f}^T \mathbf{f})$
                    
                \end{tabular}%
            }%
        }
    \end{overpic}
    \caption{Feasible force polygons for a prescribed moment of $m_z = 0$ considering different constraints on the actuator torques.}
    \label{fig:Nokleby2005-Results}
\end{figure}

The three polygons are shown in Fig. \ref{fig:Nokleby2005-Results}. The polygon obtained from considering only the maximum allowable joint torques ($||\boldsymbol{\tau}||_{\infty} \leq 4.2$) and the one obtained from the minimum-norm solution for $\boldsymbol{\tau}$ ($\min(\boldsymbol{\tau}^T \boldsymbol{\tau})$) obtained with $(\mb{J}^T \mb{K}^{-T})^{\dagger}$ are both consistent with the results reported in \cite{Nokleby2005}. The new polygon obtained using our generalized method to obtain the minimum-norm solution for $\mathbf{f}$ ($\min (\mb{f}^T \mb{f})$) differs from the latter, which the authors claimed did not produce interaction forces. 


\subsubsection{Verification}
To prove that the results obtained using (\ref{eq:EquilibratingJointTorques}) are correct, we can plot the forces generated by the solution vector $\boldsymbol{\tau}_e$. For an adequate comparison, we select a wrench vector $\mathbf{h}_o$ that is on the boundary of the two polygons that correspond to the $\text{min}(\boldsymbol{\tau}^T \boldsymbol{\tau})$ and the $\text{min}(\mathbf{f}^T \mathbf{f})$ solutions. These polygons intersect at 12 points, we choose vector $\mathbf{h}_o = \begin{bmatrix}
    1.662 & 70.689 &  0
\end{bmatrix}^T$. It is identified by the dot on Fig. \ref{fig:Nokleby2005-Results}. Solving for the joint torques using both $(\mb{J}^T \mb{K}^{-T})^{\dagger}$ and $(\mb{J}^T \mb{K}^{-T})^{\dagger}_{\mathbf{W}_e}$ yields

\begin{align}
    \boldsymbol{\tau}_{min} = (\mb{J}^T \mb{K}^{-T})^{\dagger} \mb{h}_o &= \begin{bmatrix}
        2.290 \\ 1.895 \\ -4.200 \\ 1.747 \\ 1.909 \\ -3.641
    \end{bmatrix}, \\
    \boldsymbol{\tau}_e = (\mb{J}^T \mb{K}^{-T})^{\dagger}_{\mathbf{W}_e} \mb{h}_o &= \begin{bmatrix}
        3.486 \\ 3.954 \\ -3.583 \\ 0.246 \\ 0.096 \\ -4.200
    \end{bmatrix}.
\end{align}

\noindent
We then perform the forward force problem to find the forces that these joint torque vectors generate. We obtain

\begin{equation}
    \mathbf{f} = \mb{J}^T \mb{K}^{-T} \boldsymbol{\tau}_{min} =\begin{bmatrix}
        \mathbf{f}_1 \\ \mathbf{f}_2 \\ \mathbf{f}_3
    \end{bmatrix} = \begin{bmatrix}
        -3.008 \\ 13.528 \\ -6.354 \\ 31.667 \\ 11.024 \\ 25.494
    \end{bmatrix}, 
\end{equation}

\noindent
and

\begin{equation}
    \mathbf{f}_e = \mb{J}^T \mb{K}^{-T} \boldsymbol{\tau}_e = \begin{bmatrix}
        \mathbf{f}_{e,1} \\ \mathbf{f}_{e,2} \\ \mathbf{f}_{e,3}
    \end{bmatrix} = \begin{bmatrix}
        0.554 \\ 23.563 \\ 0.554 \\ 23.563 \\ 0.554 \\ 23.563
    \end{bmatrix}.
    \label{eq:equilibratingForceDistribution1}
\end{equation}

\begin{figure}
    \centering
    \includegraphics[width=0.7\linewidth]{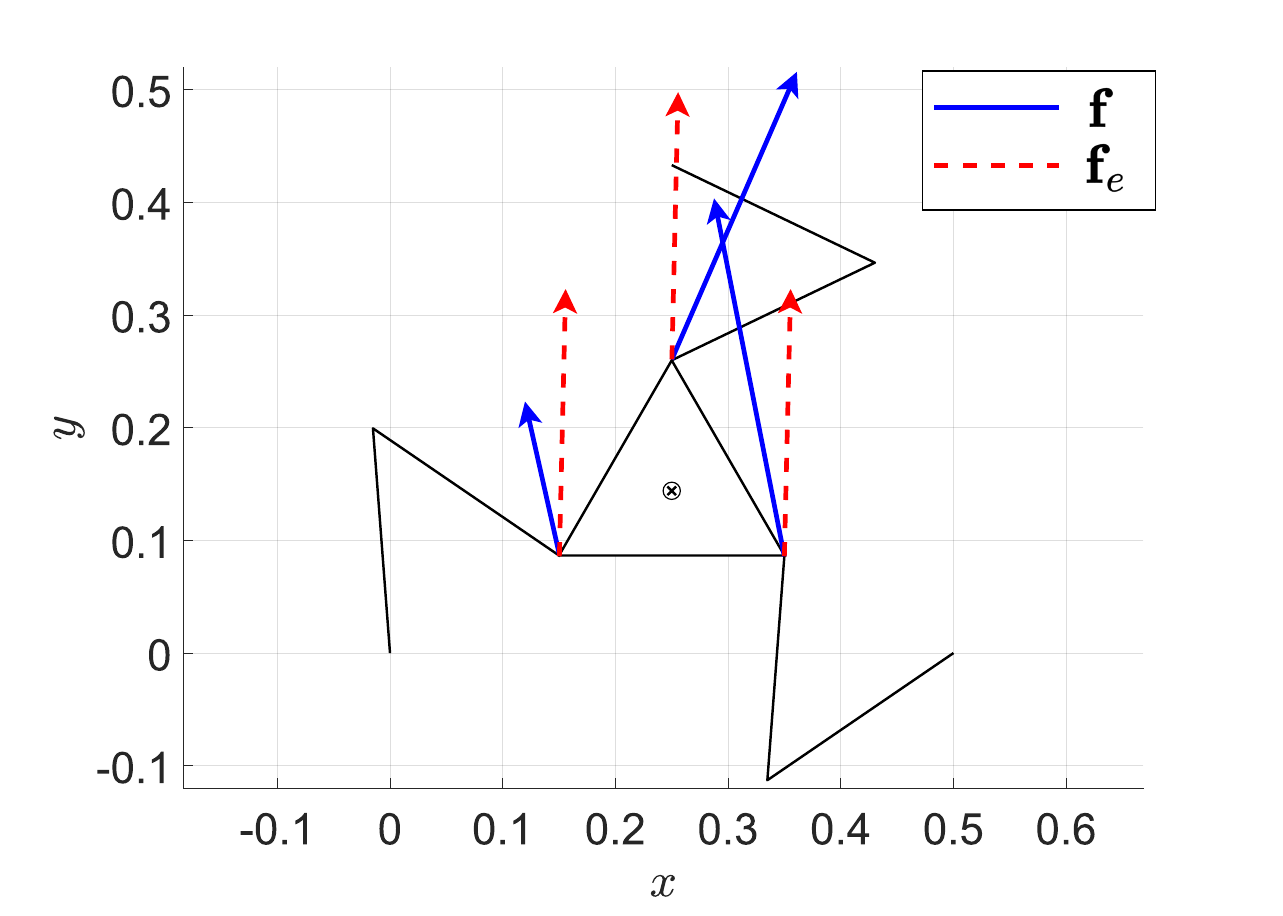}
    \caption{The force distribution $\mathbf{f}$ from \cite{Nokleby2005} and the equilibrating force distribution $\mathbf{f}_e$ which balance $\mathbf{h}_o = \begin{bmatrix}
        1.662 & 70.689 & 0
    \end{bmatrix}^T$. Forces shown are scaled by a factor of 100.}
    \label{fig:Nokleby2005-ResultantForces}
\end{figure}

\noindent
Finally, we plot these systems of forces at their respective application points. The results are shown in Fig. \ref{fig:Nokleby2005-ResultantForces}.

It is then clear that only the force system found with the weighted pseudo-inverse $(\mb{J}^T \mb{K}^{-T})^{\dagger}_{\mathbf{W}_e}$ respects the zero interaction force condition for every pair of forces (see Eq. (\ref{eq:KWInteractionForce})). 

The scaling factor method was used for similar analyses in \cite{Zibil2007, Garg2009}. We suspect that the analyses performed in these works---where the minimum-norm solution for $\boldsymbol{\tau}$ is said to not produce interaction forces---suffer from the same issues as \cite{Nokleby2005} which we have identified in this section.

\subsubsection{Effect of Changing the End-Effector Geometry}
The end-effector in the shape of an equilateral triangle used in \cite{Nokleby2005} made it so that only homogeneous virtual mass distributions $(m^*_1 = m^*_2 = m^*_3)$ were valid solutions to constraints (\ref{eq:virtualMassSum})-(\ref{eq:inertiaProportionality}). Let us now modify the end-effector geometry such that the virtual mass distribution is no longer homogeneous causing the equilibrating force distribution and the manipulating wrench distribution to differ.

For the same pose of $\mb{x} = \begin{bmatrix}
    0.250 & 0.144 & 0
\end{bmatrix}^T$, we choose to reduce the height of the triangular end-effector by half, but assume that the new end-effector is non-isotropic such that the locations of the CoM and points $C_1$ and $C_2$ remain unchanged. In other words, we assume that the CoM of the new end-effector is in the same location as the original end-effector instead of at its geometric center. The attachment point of leg 3 is also lowered so as to keep the configuration of leg 3 the same, and thus the inverse kinematic solution remains unchanged. This has no effect on the solution for the forces applied to the end-effector.

For this example, we seek to determine the equilibrating force distribution and manipulating wrench distribution for wrench $\mb{h}_o = \begin{bmatrix}
    -25 & 25 & -2
\end{bmatrix}^T$. Using (\ref{eq:EquilibratingJointTorques}), we find the equilibrating joint torque vector and the corresponding force vector that they generate: 

\begin{equation}
    \boldsymbol{\tau}_e = \begin{bmatrix}
        2.867 \\ 1.114 \\ 0.367 \\ 2.005 \\ -0.932 \\ -1.968
    \end{bmatrix} \qquad 
    \mb{f}_e = \begin{bmatrix}
        -9.810 \\ 13.447 \\ -9.810 \\ 3.220 \\ -5.381 \\ 8.333
    \end{bmatrix}
    \label{eq:equilibratingJointTorquesandForces}
\end{equation}

Then, we set $m^*_o = 3$ kg and use (\ref{eq:ManipulatingJointTorques}) to find the manipulating joint torque vector. The LMIE virtual masses that satisfy constraints (\ref{eq:virtualMassSum})-(\ref{eq:inertiaTensorEquivalence}) are $m^*_1 = m^*_2 = 0.5$ and $m^*_3 = 2$. We obtain

\begin{equation}
    \boldsymbol{\tau}_m = \begin{bmatrix}
        2.319 \\ 0.885 \\ 1.069 \\ 1.561 \\ -1.554 \\ -3.781
    \end{bmatrix} \qquad 
    \mb{f}_m = \begin{bmatrix}
        -8.016 \\ 10.834 \\ -8.016 \\ -2.501 \\ -8.969 \\ 16.667
    \end{bmatrix}
    \label{eq:manipulatingJointTorquesandForces}
\end{equation}

\begin{figure}
    \centering
    \includegraphics[width=0.7\linewidth]{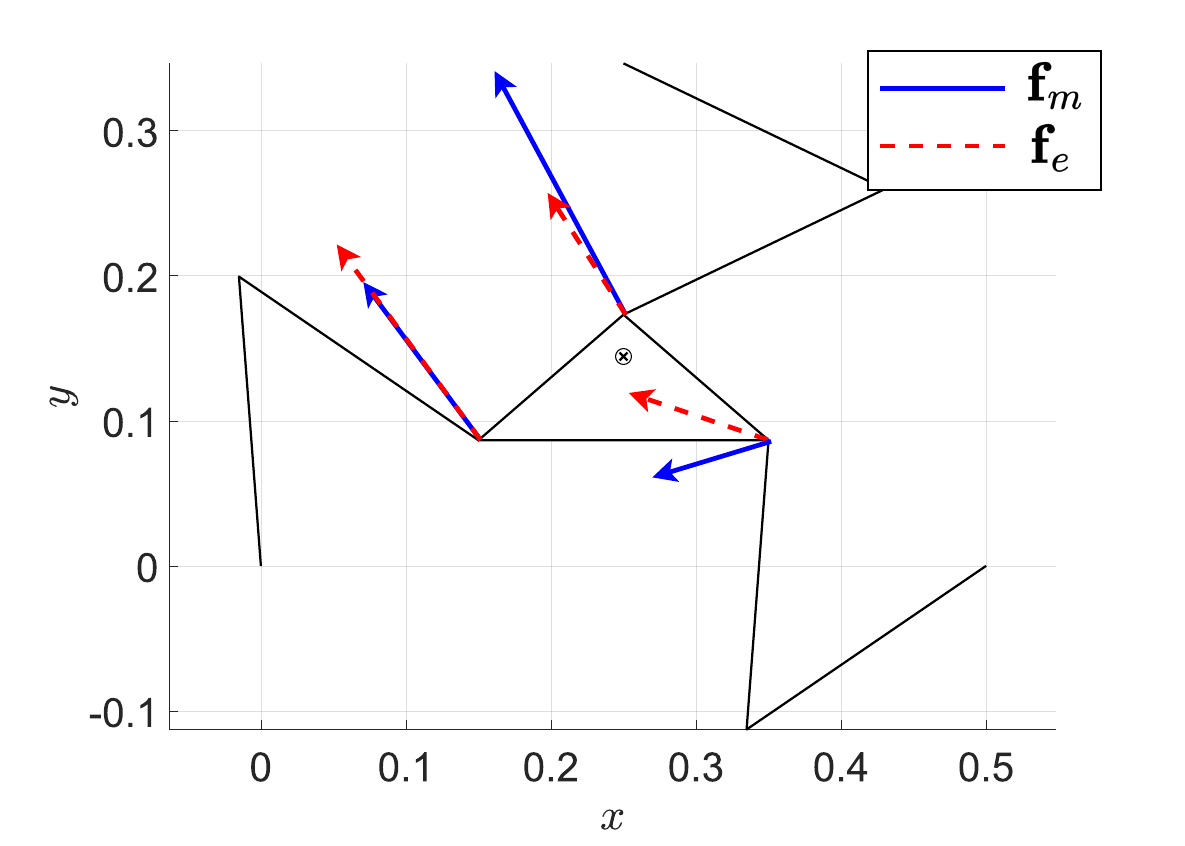}
    \caption{The equilibrating force distribution $\mathbf{f}_e$ and the manipulating force distribution $\mathbf{f}_m$ which generate $\mathbf{h}_o = \begin{bmatrix}
        -25 & 25 & -2
    \end{bmatrix}^T$. Forces shown are scaled by a factor of 100.}
    \label{fig:DiffEE-ResultantForces}
\end{figure}

Each pair of forces that can be formed from the solution in (\ref{eq:equilibratingJointTorquesandForces}) satisfies the geometric condition for no interaction forces (\ref{eq:KWInteractionForce}), and each force from the solution in (\ref{eq:manipulatingJointTorquesandForces}) induces the required constrained acceleration $\mb{x}^*_i$ of the LMIE of mass $m^*_i$ which satisfies the kinematic constraints of rigid-body motion. Both force sets are shown in Fig. \ref{fig:DiffEE-ResultantForces}. 

\subsection{Case Study 2: Spatial Manipulator With Three Serial 6-DOF Legs}
Here, we consider the most general case of a spatial manipulator with legs capable of transmitting both forces and torques to the end-effector. Architectures such as this one are seldom seen in the study of parallel manipulators, and it more closely resembles what may be encountered in cooperative manipulation tasks, but it nevertheless is practical for demonstration purposes. 

\subsubsection{Manipulator Architecture}
Consider three 6-DOF serial legs on a parallel manipulator where each leg is rigidly attached to the end-effector. The base attachment points of the legs are set as

\begin{equation}
    \mb{a}_1 = \begin{bmatrix}
        0.80 \\ 0.00  \\ 0.00 
    \end{bmatrix} \text{m} \quad
    \mb{a}_2 = \begin{bmatrix}
        -0.40 \\  0.69 \\ 0.00 
    \end{bmatrix} \text{m} \quad 
    \mb{a}_3 = \begin{bmatrix}
         -0.40 \\  -0.69 \\  0.00
    \end{bmatrix} \text{m}
\end{equation}
and the end-effector attachment points with respect to the CoM, defined in the local frame, are set as

\begin{equation}
    \mb{r}'_1 = \begin{bmatrix}
        0.25 \\ 0.00  \\ 0.00 
    \end{bmatrix} \text{m} \quad
    \mb{r}'_2 = \begin{bmatrix}
        -0.125 \\  0.217 \\ 0.00 
    \end{bmatrix} \text{m} \quad 
    \mb{r}'_3 = \begin{bmatrix}
         -0.125 \\  -0.217 \\  0.00
    \end{bmatrix} \text{m}.
\end{equation}
These can then easily be expressed in the world frame with 

\begin{equation}
    \mb{r}_i = \mb{Q} \mb{r}'_i
\end{equation}
where $\mb{Q}$ is a rotation matrix that represents the rotation from the world frame to the end-effector frame.

\begin{figure}
    \centering
    \includegraphics[width=0.7\linewidth]{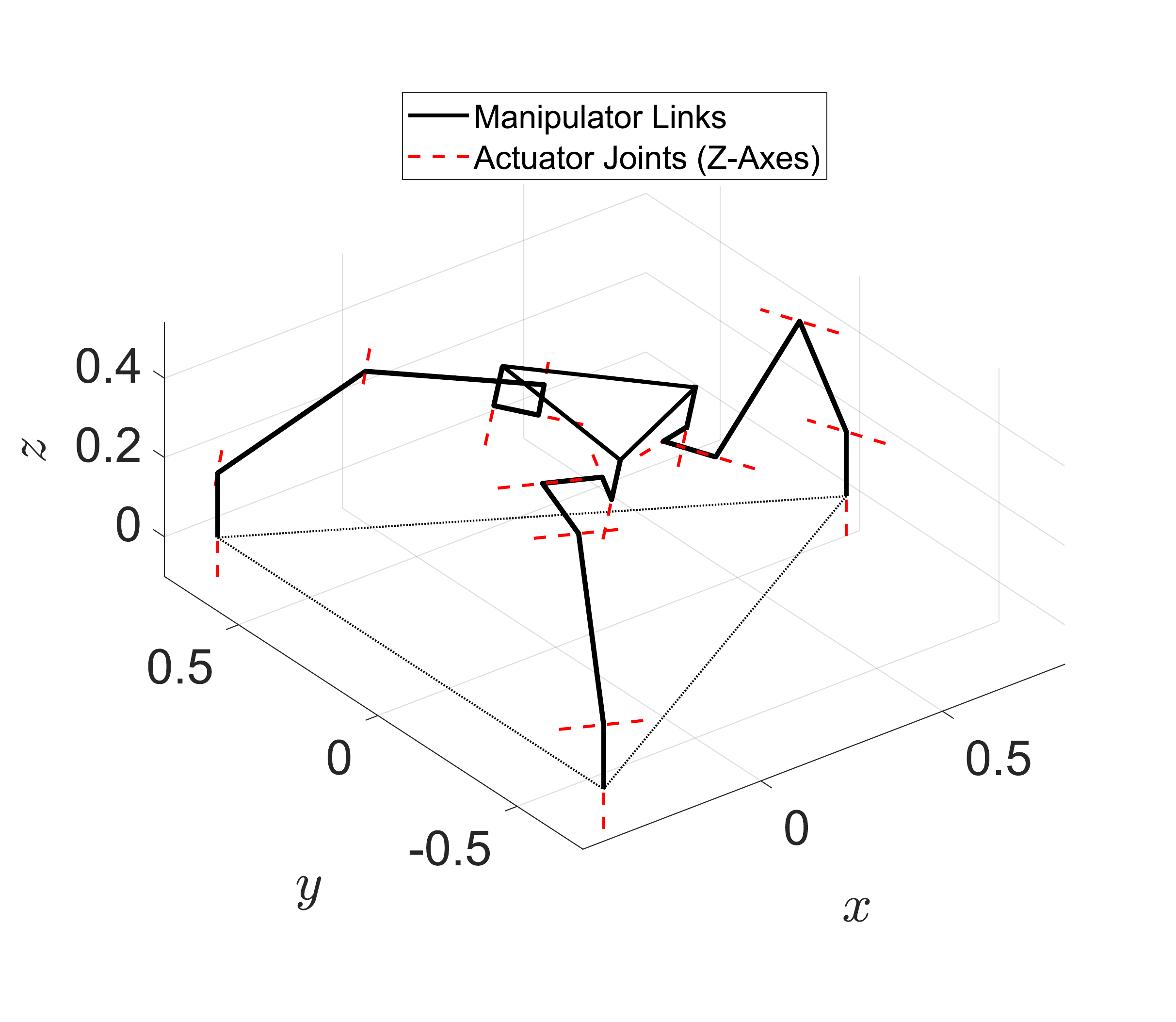}
    \caption{6-DOF spatial manipulator with three 6-DOF serial legs.}
    \label{fig:SpatialManipSchematic}
\end{figure}

{\color{black} The kinematic model of each leg is defined using the standard Denavit-Hartenberg parameters of the UR5e robot \cite{universal_robots_dh_2018}. In this formulation, the zero configuration ($\boldsymbol{\theta} = \mb{0}$) corresponds to the manipulator being fully extended horizontally along its local $x$-axis. This differs from the home position set by the manufacturer which includes an offset of $\pi/2$ on joints 2 and 4 and corresponds to the manipulator being extended vertically.} Fig. \ref{fig:SpatialManipSchematic} shows a schematic of the parallel manipulator at the chosen pose of $\mb{x}_o = \begin{bmatrix}
    \mb{p}_o^T & \mb{q}_o^T
\end{bmatrix}^T = \begin{bmatrix}
    0.1 & -0.05 & 0.5 & 5 & 10 & 0
\end{bmatrix}^T$ where the components of vector $\mb{q}_o$ define the roll, pitch, and yaw angles (in degrees) of the end-effector, in that order. {\color{black} The joint coordinates of the legs at the configuration shown in Fig. \ref{fig:SpatialManipSchematic} are

\begin{equation}
\begin{aligned}
    \boldsymbol{\theta}_1 = \begin{bmatrix}
        0.220 \\ -2.325 \\ -1.205 \\ 2.111 \\ 1.694 \\ -1.790
    \end{bmatrix}& \ \text{rad} \quad  \boldsymbol{\theta}_2 = \begin{bmatrix}
        0.239 \\ -2.198 \\ -1.288 \\ 1.731 \\ 1.636 \\ 2.366
    \end{bmatrix} \ \text{rad} \\
    \boldsymbol{\theta}_3 &= \begin{bmatrix}
        0.283 \\ -2.242 \\ -1.252 \\ 1.968 \\ 1.381 \\ 0.2286
    \end{bmatrix} \ \text{rad}.
\end{aligned}
\end{equation}}

\subsubsection{Methodology}
The static mapping of each individual leg is written as 

\begin{equation}
    \mb{J}^T_i \mb{h}_i = \boldsymbol{\tau}_i
\end{equation}
where $\mb{J}_i$ is the kinematic Jacobian of the $i$-th leg, $\mb{h}_i$ is the wrench it applies to the end-effector, and $\boldsymbol{\tau}_i$ is its vector of joint torques. To apply our methods, we first need the modified form of the static mapping as shown in  (\ref{eq:jointTorque2AppliedWrench}). When we are analyzing internal loads, we may consider that the manipulators can apply both forces and torques to the end-effector. We therefore have legs with 6 actuators generating applied wrenches with 6 components. The required matrices are written as

\begin{equation}
    \mb{K} = \mb{1}_{18} \quad \text{and} \quad \mb{B} = \begin{bmatrix}
        \mb{J}_1^{-T} & \mb{0}_{6 \times 6} & \mb{0}_{6 \times 6} \\
        \mb{0}_{6 \times 6} & \mb{J}^{-T}_2 & \mb{0}_{6 \times 6} \\
        \mb{0}_{6 \times 6} & \mb{0}_{6 \times 6} & \mb{J}^{-T}_3
    \end{bmatrix}
    \label{eq:SpatialManipKandB-Full}
\end{equation}
However, for interaction forces, we are limited to only considering pure forces. The matrices required for the modified static mapping are then written as

\begin{equation}
    \mb{K} = \mb{1}_{18} \quad \text{and} \quad \mb{B} = \begin{bmatrix}
        (\mb{J}_{1}^{-T})_f & \mb{0}_{3 \times 6} & \mb{0}_{3 \times 6} \\
        \mb{0}_{3 \times 6} & (\mb{J}^{-T}_{2})_f & \mb{0}_{3 \times 6} \\
        \mb{0}_{3 \times 6} & \mb{0}_{3 \times 6} & (\mb{J}^{-T}_{3})_f
    \end{bmatrix}
    \label{eq:SpatialManipEFD-KandB}
\end{equation}
where $(\mb{J}_{i}^{-T})_f$ is the dimensionally homogeneous part of matrix $\mb{J}_{i}^{-T}$ which maps the joint torques to the resultant force produced at the end-effector of the $i$-th manipulator. These matrices also allow us to write the static mapping of the whole manipulator as

\begin{equation}
    \mb{h}_o = \mb{G} \mb{B} \boldsymbol{\tau}
\end{equation}
where $\mb{G}$ is the grasp matrix. If $\mb{B}$ has the form shown in (\ref{eq:SpatialManipKandB-Full}), $\mb{G}$ must have the form shown in (\ref{eq:GForcesandMoments}), whereas if $\mb{B}$ has the form shown in (\ref{eq:SpatialManipEFD-KandB}), $\mb{G}$ must have the form shown in (\ref{eq:GPureForces}).

We arbitrarily set the desired output wrench to $\mb{h}_o = \begin{bmatrix}
    \mb{f}_o^T & \mb{t}_o^T
\end{bmatrix}^T = \begin{bmatrix}
    -20 & 50 & 10 & 5 & -2 & 8
\end{bmatrix}^T$. Our aim is to synthesize the vectors of joint torques which will generate this wrench while first eliminating interaction forces, then while eliminating internal loads. For the former case, the manipulators are constrained to only apply pure forces to the end-effector to remain consistent with the definition of interaction forces. For the latter case, infinitely many solutions exist for the manipulating wrench distribution. We will provide only one solution here. 

First considering the interaction force analysis, the vector of joint torques required to generate the desired wrench can be synthesized directly by using (\ref{eq:EquilibratingJointTorques}). The weighting matrix is $\mb{W}_e = \mb{B}^T \mb{B}$ with $\mb{B}$ having the form shown in (\ref{eq:SpatialManipEFD-KandB}). Note that matrix $\mb{W}_e$ is rank deficient (rank 9). Therefore, it must be inverted using singular value decomposition instead of taking the standard inverse, even though it is square. 

\subsubsection{Results}
The equilibrating joint torque vectors calculated directly with our method are

\begin{equation}
\begin{aligned}
    \boldsymbol{\tau}_{e,1} = \begin{bmatrix}
        -13.09 \\ 4.74 \\ 1.45 \\ -0.44 \\ -2.92 \\ 0
    \end{bmatrix} \text{Nm}& \quad \boldsymbol{\tau}_{e,2} = \begin{bmatrix}
        -4.42 \\ 13.99 \\ 5.35 \\ 1.27 \\ -0.36 \\ 0
    \end{bmatrix} \text{Nm} \\ \boldsymbol{\tau}_{e,3} = & \begin{bmatrix}
        3.63 \\ -8.50 \\ -2.51 \\ -0.23 \\ 0.18 \\ 0
    \end{bmatrix} \text{Nm}.
\end{aligned}
\end{equation}

To verify this solution, we can calculate the force vectors produced by these joint torque vectors. We obtain

\begin{equation}
\begin{aligned}
    \mb{f}_{e,1} = \begin{bmatrix}
        -5.92 \\ 28.19 \\ 7.57
    \end{bmatrix} \text{N} \ \mb{f}_{e,2} = \begin{bmatrix}
        -15.73 \\ 10.20 \\ 11.01
    \end{bmatrix} \text{N} \
    \mb{f}_{e,3} = \begin{bmatrix}
        1.65 \\ 11.62 \\ -8.58
    \end{bmatrix} \text{N}
\end{aligned}
\end{equation}
where it can easily be verified that this force distribution satisfies constraint (\ref{eq:KWInteractionForce}) and is therefore the equilibrating force distribution.

We proceed similarly to determine the manipulating torque vectors, although the solution is not unique in this case, since we may choose which portion of the resultant torque $\mb{t}_o$ is assigned to the manipulating forces $\mb{f}_{m,i}$ and which portion is assigned to the manipulating torques $\mb{t}_{m,i}$ applied by the legs. In this example, we consider the case where all of $\mb{t}_o$ is assigned to the manipulating torques $\mb{t}_{m,i}$, since it is easier to verify. 


Synthesizing the manipulating torque vector $\boldsymbol{\tau}_m$ involves calculating the virtual mass $m^*_i$ and a virtual inertia tensor $\mb{I}^*_i$ of a set of three LMIEs located at the three leg attachment points. As it was pointed out in \cite{Flight2026}, when the attachment points of the kinematic chains to the rigid body are pre-defined, a total of 10 forces are required to completely eliminate internal loads. However, when the center of mass of the rigid body lies within the convex hull defined by the leg attachment points, a simplified (and less restrictive) solution procedure can be used. This is done by not considering constraint (\ref{eq:inertiaProportionality}), which is most restrictive, and solving the reduced linear system defined by 

\begin{equation}
    \mb{R} \mb{m}^* = \begin{bmatrix}
        1 & 1 & 1 \\ \mb{r}'_1 & \mb{r}'_2 & \mb{r}'_3
    \end{bmatrix} \begin{bmatrix}
        m^*_1 \\ m^*_2 \\ m^*_3
    \end{bmatrix} =  \begin{bmatrix}
        m^*_o \\ 0 \\ 0 \\ 0
    \end{bmatrix}
\end{equation}
where $\mb{r}'_i \in \mathbb{R}^2$ defines the location of the attachment point of the $i$-th leg in the plane of the end-effector in the local frame. {\color{black} Setting $m^*_o = 3$ kg and} solving this for our current example, we obtain $\mb{m}^* = \begin{bmatrix}
    1 & 1 & 1
\end{bmatrix}$. 

This enables us to write the inertia matrix of the system as

\begin{equation}
    \mb{M} = \begin{bmatrix}
        \mb{M}^*_1 & \mb{0}_{6 \times 6} & \mb{0}_{6 \times 6} \\
        \mb{0}_{6 \times 6} & \mb{M}^*_2 & \mb{0}_{6 \times 6} \\
        \mb{0}_{6 \times 6} & \mb{0}_{6 \times 6} & \mb{M}^*_3
    \end{bmatrix}
\end{equation}
where $\mb{M^*}_i$ is the inertia matrix of the $i$-the LMIE written as 

\begin{equation}
    \mb{M}^*_i = \begin{bmatrix}
        m^*_i \mb{1}_3 & \mb{0}_{3 \times 3} \\
        \mb{0}_{3 \times 3} & \mb{I}^*_i
    \end{bmatrix}.
\end{equation}

The manipulating joint torque vector can now be synthesized using (\ref{eq:ManipulatingJointTorques}), and different solutions can be obtained by choosing different values for the inertial parameters $(m^*_i, \mb{I}^*_i)$ of the LMIEs that parametrize the solution space.


        
        
        


        
        
        

To weight the resultant torque entirely on the manipulating torques, we assign large values to virtual inertias of the LMIEs such that their contribution to the total inertia of the system is much larger than that of the point masses $m^*_i$ (i.e., $\sum_{i=1}^n \mb{I}^*_i >> \sum_{i=1}^n \mb{S}(\mb{r}_i) m^*_i \mb{S}(\mb{r}_i)^T$). Note that we are still using the solution $\mb{m}^* = \begin{bmatrix}
    1 & 1 & 1
\end{bmatrix}$ for the virtual masses. For the inertia tensors of the LMIEs, we choose

\begin{equation}
    \mb{I}^*_i = 10^6 \sum_{i=1}^n \mb{S}(\mb{r}_i) m^*_i \mb{S}(\mb{r}_i)^T \quad \text{for} \ i = 1, 2, 3.
\end{equation}
This ensures that the inertia contribution from the inertia tensors of the LMIEs dominates that of the point masses while also satisfying the proportionality constraint (\ref{eq:inertiaProportionality}). Such large values are permitted since, again, the inertial properties $(m^*_o, \mb{I}^*_o)$ of the dynamically equivalent system of LMIEs need only be proportional to those of the end-effector $(m_o, \mb{I}_o)$ (see Section \ref{sec:Def-InternalLoads} and Ref. \cite{Flight2026}). The manipulating joint torque vector is

\begin{equation}
\begin{aligned}
    \boldsymbol{\tau}_{m,1} = \begin{bmatrix}
        -5.23 \\ 0.20 \\ -0.41 \\ -1.61 \\ -0.87 \\ 2.96
    \end{bmatrix} \text{Nm}& \ 
    \boldsymbol{\tau}_{m,2} = \begin{bmatrix}
        5.49 \\ 7.61 \\ 2.08 \\ 0.98 \\ -0.39 \\ 2.96
    \end{bmatrix} \text{Nm} \\
    \boldsymbol{\tau}_{m,3} = &\begin{bmatrix}
        11.74 \\ -0.53 \\ 2.63 \\ 0.27 \\ 1.23 \\ 2.96
    \end{bmatrix} \text{Nm}.
\end{aligned}
\end{equation}

Again, we can easily calculate the wrenches generated by these joint torque vectors. We obtain

\begin{equation}
    \mb{h}_{m,1} = \begin{bmatrix}
        -6.67 \\ 16.67 \\ 3.33 \\ 1.67 \\ -0.67 \\ 2.67 
    \end{bmatrix} \
    \mb{h}_{m,2} = \begin{bmatrix}
        -6.67 \\ 16.67 \\ 3.33 \\ 1.67 \\ -0.67 \\ 2.67 
    \end{bmatrix} \ 
    \mb{h}_{m,3} = \begin{bmatrix}
        -6.67 \\ 16.67 \\ 3.33 \\ 1.67 \\ -0.67 \\ 2.67 
    \end{bmatrix}.
    \label{eq:ManipulatingWrenchDist2}
\end{equation}

This last solution is particularly interesting since it is easy to validate that the wrenches produced by the manipulating joint torque vector obtained with the proposed method correspond to a valid manipulating wrench distribution. Since all of the resultant torque $\mb{t}_o$ was assigned to the manipulating torques $\mb{t}_{m,i}$, the manipulating forces $\mb{f}_{m,i}$ need only generate the resultant force $\mb{f}_o$. This is equivalent to a pure force being applied to the centroid of the point mass system, and every particle in the system therefore has the same instantaneous acceleration $\ddot{\mb{p}}^*_i$. The manipulating forces are obtained from 

\begin{equation}
    \mb{f}_{m,i} = m^*_i \ddot{\mb{p}}^*_i,
\end{equation}
and we see that each manipulating force is weighted by the virtual mass of the LMIE to which it is applied. For this manipulator, we have a homogeneous mass distribution ($m^*_1 = m^*_2 = m^*_3$). The expected result is therefore that the manipulating forces should be parallel, be of equal magnitude, and sum to the resultant force $\mb{f}_o$, which is what we observe in the solution shown in (\ref{eq:ManipulatingWrenchDist2}).

As for the manipulating torques, we know that they should be parallel and sum to the resultant torque $\mb{t}_o$. Since we have decided to weight the virtual inertia tensors equally ($\mb{I}^*_1 = \mb{I}^*_2 = \mb{I}^*_3$), we also know that they should be of equal magnitude. The manipulating torques shown in (\ref{eq:ManipulatingWrenchDist2}) are consistent with these constraints.

\section{Conclusion}
The aim of this paper is to adapt existing null-space wrench component analysis techniques for grasp-like systems to parallel manipulators with actuation redundancy. We first review existing adaptations and demonstrate how they fail to account for the fundamental differences in the wrench mappings of these two types of systems. We then provide our own adaptations by deriving generalized methods which do take into consideration the nature of parallel manipulator wrench transmission. 

Two novel solutions to the wrench synthesis problem for redundant parallel manipulators are presented. The first is a corrected form of an existing method for synthesizing equilibrating joint torque vectors. Our generalized method properly incorporates the mapping of joint torques to applied wrenches into a weighted Moore-Penrose pseudo-inverse of the wrench mapping defined by $\mb{J}^T \mb{K}^{-T}$. We prove the correctness of our approach {\color{black}with a comparative analysis. The results show that only the wrench vectors found with our method satisfy the necessary geometric conditions for eliminating interaction forces while the others do not. Our second proposed method is a first of its kind and allows for the synthesis of manipulating joint torque vectors which do not generate internal loads. The differences between the equilibrating force distribution and the manipulating force distribution are shown mathematically and through detailed case studies. }

The results obtained from our analyses provide two major insights concerning wrench distribution in parallel manipulators with actuation redundancy. First, interaction forces and internal loads are distinct physical phenomena and must be treated as such. We have clearly defined both in this paper in the hopes that they will no longer be conflated in the robotics literature. Second, the wrench mapping of parallel manipulators differs from that of grasp-like systems, and this difference must be considered when adapting methods for the analysis of one system to the other. 

We are confident that the explicit methods for joint torque synthesis proposed in this paper will prove useful for manipulator design, analysis, and control, since estimating the wrench capabilities of a manipulator and intelligently prescribing joint torques during path planning are critical for these tasks. Future work should notably be aimed at adapting our novel methods to wrench capability analysis of parallel manipulators and real-time torque control algorithms.

\section*{Acknowledgments}
The authors would like to acknowledge the financial support of the Natural Sciences and Engineering Research Council of Canada (NSERC) through a scholarship (Canada Graduate Research Scholarship — Doctoral) to the first author and discovery grant DG-89715 to the second author. They would also like to acknowledge the financial support of the Institut de recherche Robert-Sauvé en santé et sécurité du travail (IRSST) through a doctoral scholarship to the first author. 






\appendix

\section{Velocity Equations of the 3-\underline{RR}R Planar Parallel Manipulator}
Let $\mb{p} = \begin{bmatrix}
    x & y
\end{bmatrix}^T$ define the position of the center of mass of the end-effector in the base frame. The loop closure equation of the $i$-th leg is then written as 

\begin{equation}
    \mb{p} + \mb{r}_i = \mb{a}_i + \mb{u}_i + \mb{v}_i.
    \label{eq:LoopClosure}
\end{equation}
Rearranging the terms, we obtain 

\begin{equation}
    \mb{v}_i^T \mb{v}_i = (\mb{p} + \mb{r}_i - \mb{a}_i - \mb{u}_i)^T(\mb{p} + \mb{r}_i - \mb{a}_i - \mb{u}_i).
\end{equation}

The first time derivative, after simplification, can be written as

\begin{equation}
    (\mb{p} + \mb{r}_i - \mb{a}_i - \mb{u}_i)^T (\dot{\mb{p}} + \dot{\mb{r}}_i) = (\mb{p} + \mb{r}_i - \mb{a}_i - \mb{u}_i)^T \dot{\mb{u}}_i,
    \label{eq:TimeDerivative1}
\end{equation}
By substituting (\ref{eq:LoopClosure}) into (\ref{eq:TimeDerivative1}), we obtain

\begin{equation}
    \mb{v}_i^T (\dot{\mb{p}} + \dot{\mb{r}}_i) = \mb{v}_i^T \dot{\mb{u}}_i
\end{equation}

In a planar system, the time derivatives $\dot{\mb{r}}_i$ and $\dot{\mb{u}}_i$ can be written as 

\begin{align}
    \dot{\mb{r}}_i &= \mb{E} \mb{r}_i \dot{\phi}, \label{eq:rDot} \\
    \dot{\mb{u}}_i &= \mb{E} \mb{u}_i \dot{\theta}_{1i}.
\end{align}
We arrive at 

\begin{equation}
    \mb{v}_i^T \dot{\mb{p}} + \mb{v}_i^T \mb{E} \mb{r}_i \dot{\phi} = \mb{v}_i^T \mb{E} \mb{u}_i \dot{\theta}_{1i}.
    \label{eq:Joint1VelocityEquation}
\end{equation}

The velocity equation of the second actuated joint can be obtained by applying the cosine law to triangle $A_i B_i C_i$. This is written as

\begin{equation}
    (\mb{p} + \mb{r}_i - \mb{a}_i)^T (\mb{p} + \mb{r}_i - \mb{a}_i) = ||\mb{u}_i||^2 + ||\mb{v}_i||^2 - 2||\mb{u}_i||||\mb{v}_i|| \cos (\theta'_{2i})
\end{equation}
where $\theta'_{2i}$ is the inside angle between vectors $\mb{u}_i$ and $\mb{v}_i$. Taking the first time derivative, we obtain

\begin{equation}
    (\mb{p} + \mb{r}_i - \mb{a}_i)^T (\dot{\mb{p}} + \dot{\mb{r}}_i) = ||\mb{u}_i||||\mb{v}_i|| \sin (\theta'_{2i}) \dot{\theta}'_{2i}.
\end{equation}
Using (\ref{eq:rDot}) and noting that $\dot{\theta}'_{2i} = \dot{\theta}_{2i}$, this further simplifies to 

\begin{equation}
    (\mb{u}_i + \mb{v}_i)^T \dot{\mb{p}} + (\mb{u}_i + \mb{v}_i)^T \mb{E} \dot{\mb{r}}_i \dot{\phi} = ||\mb{u}_i||||\mb{v}_i|| \sin (\theta'_{2i}) \dot{\theta}_{2i}.
    \label{eq:Joint2VelocityEquation}
\end{equation}

When we gather the velocity equations (\ref{eq:Joint1VelocityEquation}) and (\ref{eq:Joint2VelocityEquation}) of the three legs in matrix form, we obtain the kinematic Jacobians $\mb{J}$ and $\mb{K}$ shown in (\ref{eq:3RRR-J}) and (\ref{eq:3RRR-K}), respectively. 

\printcredits

\bibliographystyle{cas-model2-names}

\bibliography{bibliography}



\end{document}